%% file: main.tex
\documentclass{article}

\PassOptionsToPackage{numbers, compress}{natbib}
\usepackage[sglblindworkshop, final]{neurips_2026}
\workshoptitle{Interpretability for Discovery: Understanding and Discovering Novel Knowledge in AI Models}

\usepackage[utf8]{inputenc}
\usepackage[T1]{fontenc}
\usepackage{hyperref}
\usepackage{url}
\usepackage{booktabs}
\usepackage{amsfonts}
\usepackage{nicefrac}
\usepackage{microtype}
\usepackage[table]{xcolor}

\usepackage{graphicx}
\usepackage{amsmath}
\usepackage{subcaption}
\usepackage{enumitem}
\usepackage{bm}

\usepackage[textsize=scriptsize]{todonotes}
\definecolor{rowgray}{gray}{0.92}

\title{Match and Correct: A Repeated Computational Motif in Evoformer Layers}

\author{%
  Ishan Khire$^{1}$ \quad
  Naren Manikandan$^{1}$ \quad
  Tyler Hayes$^{2}$\setcounter{footnote}{6}\thanks{Work done while at the Georgia Institute of Technology.} \quad
  Giri Krishnan$^{1}$\setcounter{footnote}{0}\thanks{Corresponding author: \texttt{giri@gatech.edu}} \\[0.5ex]
  $^{1}$College of Computing, Georgia Institute of Technology \quad
  $^{2}$Arsenale Bioyards
}

\begin{document}

\maketitle

\begin{abstract}
    Triangle attention and triangle multiplicative updates form key components of the Evoformer, and were introduced specifically to encourage consistency with the triangle inequality. Yet whether representational changes over layers reduce triangle inequality violations has not been tested. We probe the pair representation at each of OpenFold's 48 Evoformer layers, decoding final C$_\alpha$--C$_\alpha$ distances with four probe families (a shared linear probe, per-protein linear probes, a sequence-separation-gated linear probe, and XGBoost) and reconstructing intermediate 3D structures from the decoded distances via multidimensional scaling. Across a cohort of 552 CAMEO and ECOD proteins, distance decodability ($R^2$) and reconstruction quality (TM-score) rise in early layers, plateau in the middle of the network, and rise again in late layers. Directly measuring triangle-inequality violations in the decoded distances reveals a similar, more exaggerated pattern: violations rise to a peak, fall to a trough, rise to a second peak, and only then decline, revealing two distinct episodes of decreasing normalized triangle violations rather than one continuous improvement. Cumulative forward and reverse layer-wise ablations of the Pair Transition, MSA Transition, and triangle mechanisms localize the regions in which specific Evoformer components become essential. The two stages of geometric refinement coincide with the use of the Pair Transition MLP and the triangle mechanisms. These results characterize how distance information and decoded geometric consistency evolve across layers, and motivate hypotheses about the computations behind their trajectories.

\end{abstract}

\section{Introduction}

AlphaFold2, and its open-source reproduction OpenFold, can predict protein structures with near-atomic accuracy \cite{jumper2021,ahdritz2024}. Yet the success of these models leaves a fundamental question unanswered: \emph{what are the evolutionary, geometric, and physical principles that determine the protein structure, and how are these principles implemented through the network?} Understanding where they emerge within the model may provide insight into general principles of protein structure and improve AI approaches for protein sciences.

The Evoformer introduced in AlphaFold2 \cite{jumper2021} and the Pairformer introduced in AlphaFold3 \cite{abramson2024af3} use triangle attention and triangle multiplicative updates, which are considered key components enabling accurate structure prediction. These modules update the pair representation of a residue pair $i, j$ using the representations of pairs $i, k$ and $j, k$ for all $k$. The motivation is explicitly geometric: a set of pairwise distances describes a real 3D structure only if it satisfies constraints such as the triangle inequality, a classical condition from distance geometry \cite{havel1983}. Routing information between the edges of a triangle is meant to help the model respect these constraints. The Evoformer and Pairformer rely on this inductive bias, but whether the pair representation actually satisfies such constraints has not, to our knowledge, been tested directly on the model's representations. AlphaFold2 includes a distogram head that predicts distributions over pairwise distance bins from the final Evoformer pair representation.

Hence, we ask: through Evoformer layers, is there a progressive increase in the geometric consistency of structural information encoded in the pair representation?
We address this by probing the pair representation at each of OpenFold's 48 Evoformer layers and decoding C$_\alpha$--C$_\alpha$ distances with four different probes. We show that distance decodability rises in early layers, followed by a plateau in middle layers, and then an increase again. Furthermore, we use multidimensional scaling (MDS) \cite{torgerson1952} to reconstruct intermediate structures from decoded pairwise distances and find that their Template Modeling (TM) scores \cite{zhangskolnick2004} follow a similar layerwise trend.
We test whether every unordered residue triplet follows the triangle inequality, and if not, we measure the magnitude of the violation. We then compare the trajectories of these trends with the trajectories of cumulative forward and reverse ablations to see where stages of geometric refinement coincide with layers during which specific components become essential. We interpret these findings by proposing a potential ``match and correct'' hypothesis: MSA-side computations aggregate evolutionary information and pair-side computations refine geometric information.

\section{Related work}

There are two broad views on how AlphaFold2-style models arrive at a structure. One view treats the network as performing sophisticated fold recognition from evolutionary data, followed by comparatively small geometric refinements \cite{skolnick2021alphafold}. This view is consistent with evidence that structure trajectories from current structure predictors do not appear to simulate protein folding dynamics \cite{outeiral2022}. A second view holds that AlphaFold2 implicitly learns something closer to an energy function, since its confidence estimates can accurately rank candidate structures in order of similarity to a target sequence's structure \cite{roney2022}. These accounts are not mutually exclusive, and neither directly examines the pair representation's \emph{internal} geometric consistency as it evolves across depth, which is the question we study here. Ouyang-Zhang et al.~\cite{pairmixer2025} argue that pair representations are key to capturing relationships between residue pairs, and triangle multiplication is effective at refining these interactions by reasoning over residue triplets. This reinforces the idea that triangle-consistent geometry is central to why these architectures work; we complement this architectural argument with a direct, layer-by-layer measurement of triangle-inequality violations in distances decoded from a trained model's pair representation. Lu et al.~\cite{lu2026} causally probe ESMFold, OpenFold, and Boltz-1 \cite{wohlwend2024} and report a shared two-stage computational structure, in which early blocks propagate sequence-derived biochemical signals into the pair representation and late blocks accumulate spatial (distance and contact) features. Our results are consistent with a late-stage focus on spatial refinement, and further show that decoded triangle consistency improves in two separate episodes across the Evoformer. More broadly, step-by-step or staged computation has also been reported in models outside protein structure prediction: Lad et al.~\cite{lad2024} identify discrete stages of inference in large language models, and probing methods more generally are reviewed by Belinkov~\cite{belinkov2022}.

\section{Results}

\begin{figure}
    \centering
    \begin{subfigure}[t]{0.49\linewidth}
        \centering
        \includegraphics[width=\linewidth]{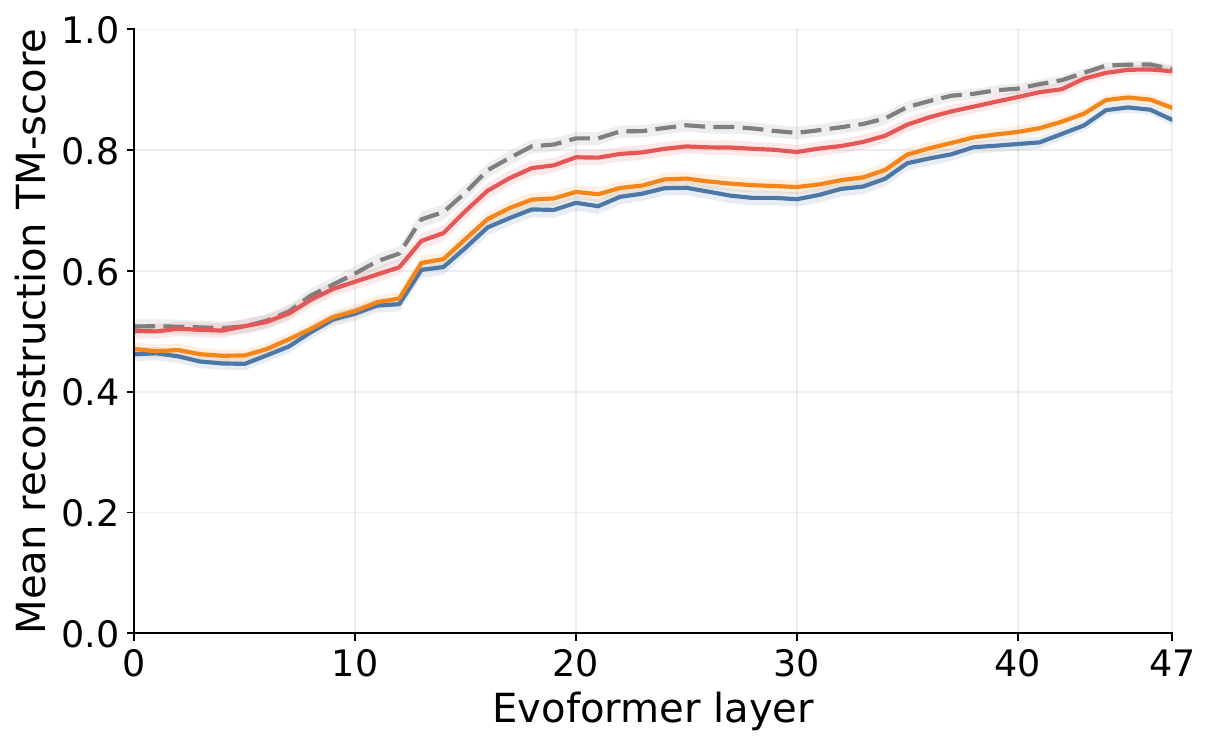}
        \caption{Mean reconstruction TM-score}
        \label{fig:tm-score}
    \end{subfigure}\hfill
    \begin{subfigure}[t]{0.49\linewidth}
        \centering
        \includegraphics[width=\linewidth]{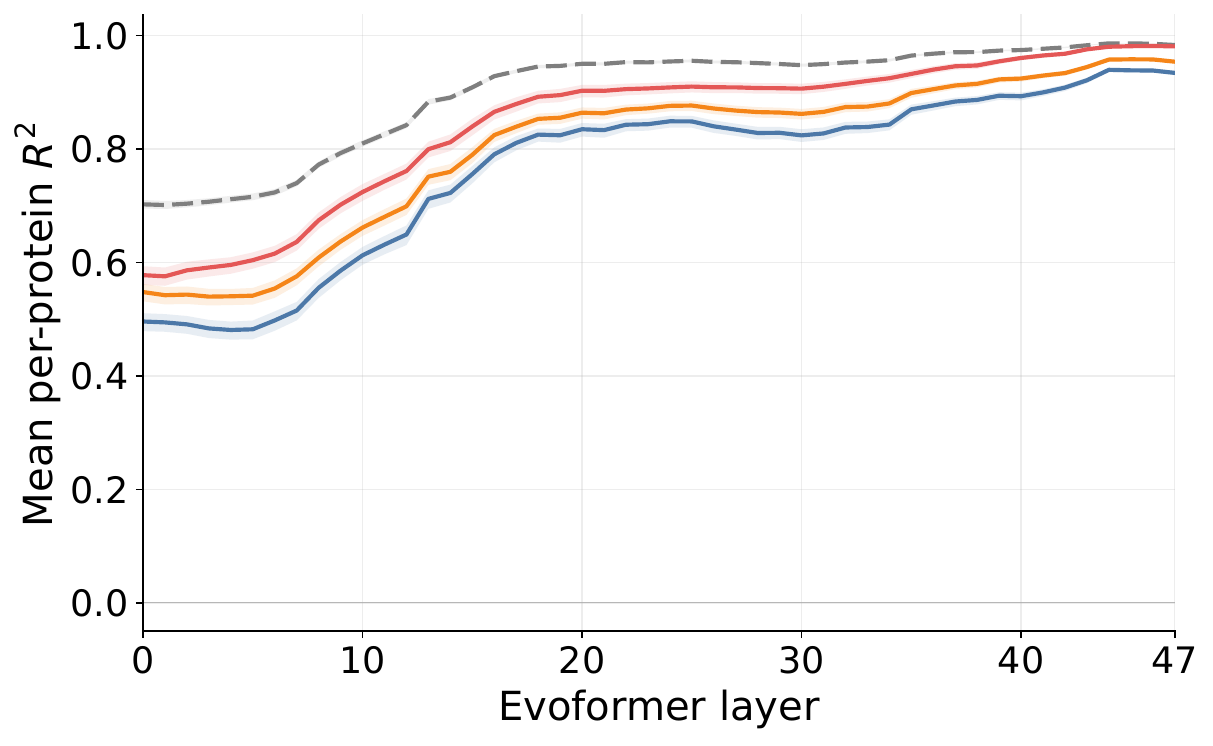}
        \caption{Mean per-protein $R^2$}
        \label{fig:mean-per-protein-r2}
    \end{subfigure}

    \vspace{0.5em}

    \begin{subfigure}[t]{\linewidth}
        \centering
        \includegraphics[width=\linewidth]{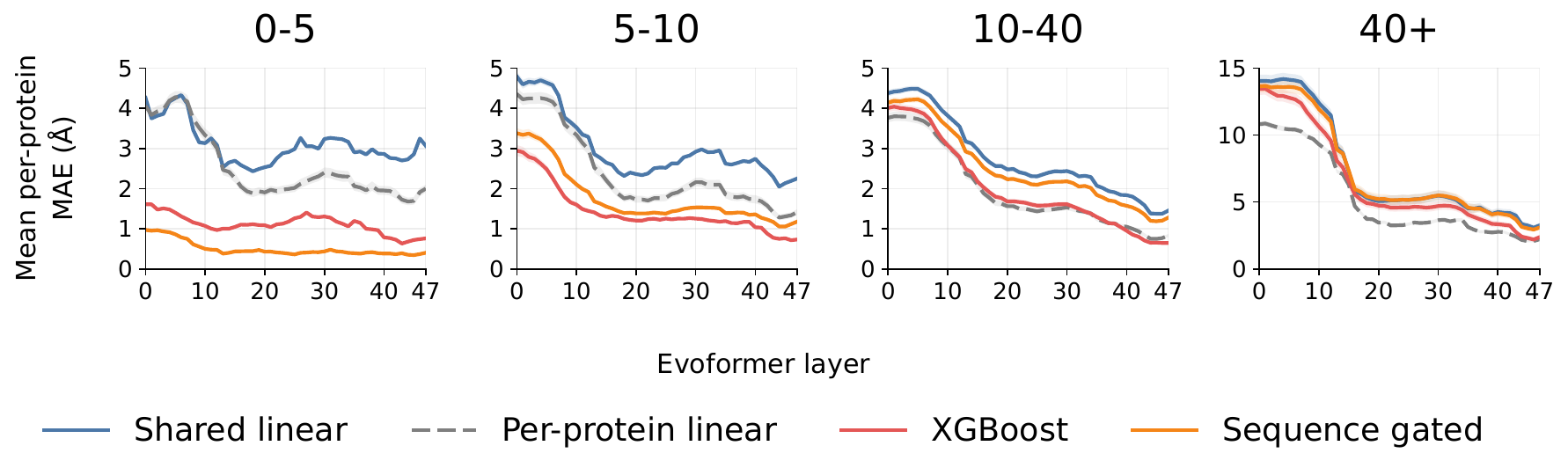}
        \caption{Distance-prediction MAE by spatial-distance bin}
        \label{fig:spatial-mae}
    \end{subfigure}
    \caption{Layerwise structural reconstruction and distance decodability across the Evoformer. Top: mean reconstruction TM-score and mean per-protein $R^2$. Bottom: mean per-protein distance-prediction MAE stratified by true spatial distance between residues (0--5, 5--10, 10--40, and $\geq40$~\AA). Curves compare shared linear, per-protein linear, XGBoost, and sequence-gated probes. Shaded bands in the MAE panels show paired-protein bootstrap 95\% confidence intervals.}
    \label{fig:decodability-and-spatial-mae}
\end{figure}

\subsection{Decoding pairwise distances}

We used OpenFold \cite{ahdritz2024}, an open-source implementation based on AlphaFold2 \cite{jumper2021}, with recycling disabled. Thus, a single forward pass through the 48 Evoformer layers constitutes the complete trunk computation. We cached the pair representation $Z^{(\ell)} \in \mathbb{R}^{R \times R \times 128}$ at each layer $\ell$, where $R$ is the number of residues in the protein. For each unordered residue pair $(i,j)$, we concatenated the two directional pair embeddings, $Z^{(\ell)}_{ij}$ and $Z^{(\ell)}_{ji}$, yielding a 256-dimensional feature vector.

At each layer $\ell$, probes were trained to predict the C$_\alpha$--C$_\alpha$ distance $d_{ij}$ between residues $i$ and $j$ in the final structure generated by OpenFold. The target structure was therefore held fixed across all layers, while only the pair representation provided to the probe varied. Comparing probe performance across layers consequently measures the extent to which information about the final pairwise geometry is decodable from the pair representation at each stage of the Evoformer. Importantly, this procedure does not require or assume that an intermediate pair representation is itself a complete structural prediction; rather, it tests whether information about the final structure is accessible from that representation.

We trained four regression models as probes at each layer: (i) a shared linear probe trained across all proteins, (ii) per-protein linear probes fitted independently to each protein, (iii) sequence-separation-gated linear probes, each fitted separately within each of six sequence-separation ($|i-j|$) bins, and (iv) a shared nonlinear XGBoost regression model \cite{chen2016}. A detailed description of the dataset and methods is provided in Supplement~\S\ref{sec:supp-methods}.

Qualitatively, the change in $R^2$ across layers indicates three phases in the Evoformer (Figure~\ref{fig:mean-per-protein-r2}). From layer 0 to approximately layer 18, $R^2$ for the shared linear probe rises from 0.50 to 0.83. From layer 18 to 33, it plateaus, changing only from 0.83 to 0.84. In the third phase, from layer 33 to 47, $R^2$ again increases from 0.84 to 0.93. The $R^2$ values of the per-protein linear probe and the XGBoost model show a similar three-phase pattern, with transitions occurring at approximately the same layers (Figure~\ref{fig:supp-individual-r2}).

\subsection{Intermediate structure prediction}

To assess whether the decoded distances captured meaningful three-dimensional structure, we reconstructed structures from the predicted pairwise distance matrices using multidimensional scaling (MDS). The resulting TM-scores provide a structural validation of the information recovered by the probes. Across all four probe families, the reconstructed structures already achieved mean per-protein TM-scores above 0.4 in the earliest layers, exceeded 0.5 by layer 9, and surpassed 0.7 by layer 18 (Figure~\ref{fig:tm-score}; individual-protein trajectories in Figure~\ref{fig:supp-individual-tm-score}). Because a TM-score above 0.5 generally indicates that two structures share the same overall fold \cite{xu2010}, these results indicate that substantial information about the global protein fold is already decodable from the pair representation in the early Evoformer layers. The TM-score trajectory also exhibits the three phases seen in $R^2$: a rise through the early layers, a plateau across the middle layers, and a second rise in the late layers (Figure~\ref{fig:tm-score}).

\subsection{Probe accuracy stratified by spatial distance}

To test whether probe error depended on the true spatial separation of a residue pair, we stratified residue pairs by their C$_\alpha$--C$_\alpha$ distance in the final structure and computed the mean per-protein absolute error (MAE) within each distance bin at every layer (Figure~\ref{fig:spatial-mae}).

From layers 1 to 20, mean per-protein MAE decreased in every spatial-distance bin for all four probes. Between layers 20 and 30, this trend reversed for short-range pairs: MAE increased consistently for 5--10~\AA{} pairs, and for 0--5~\AA{} pairs in three of the four probes. Over the same layers, MAE was roughly stagnant for 10--40~\AA{} pairs, whereas errors for $\geq40$~\AA{} pairs increased slightly for three probes and showed no clear change for XGBoost. From layers 30 to 47, MAE decreased again in every bin and for every probe family. Taken together, these results point to two stages of error correction, one ending around layer 20 and one beginning around layer 30, each reducing error at all distances, separated by an intermediate phase in which errors slightly rise for short-range pairs.

\subsection{Triangle inequality violations}

\begin{figure}
    \centering
    \begin{subfigure}[t]{0.49\linewidth}
        \centering
        \includegraphics[width=\linewidth]{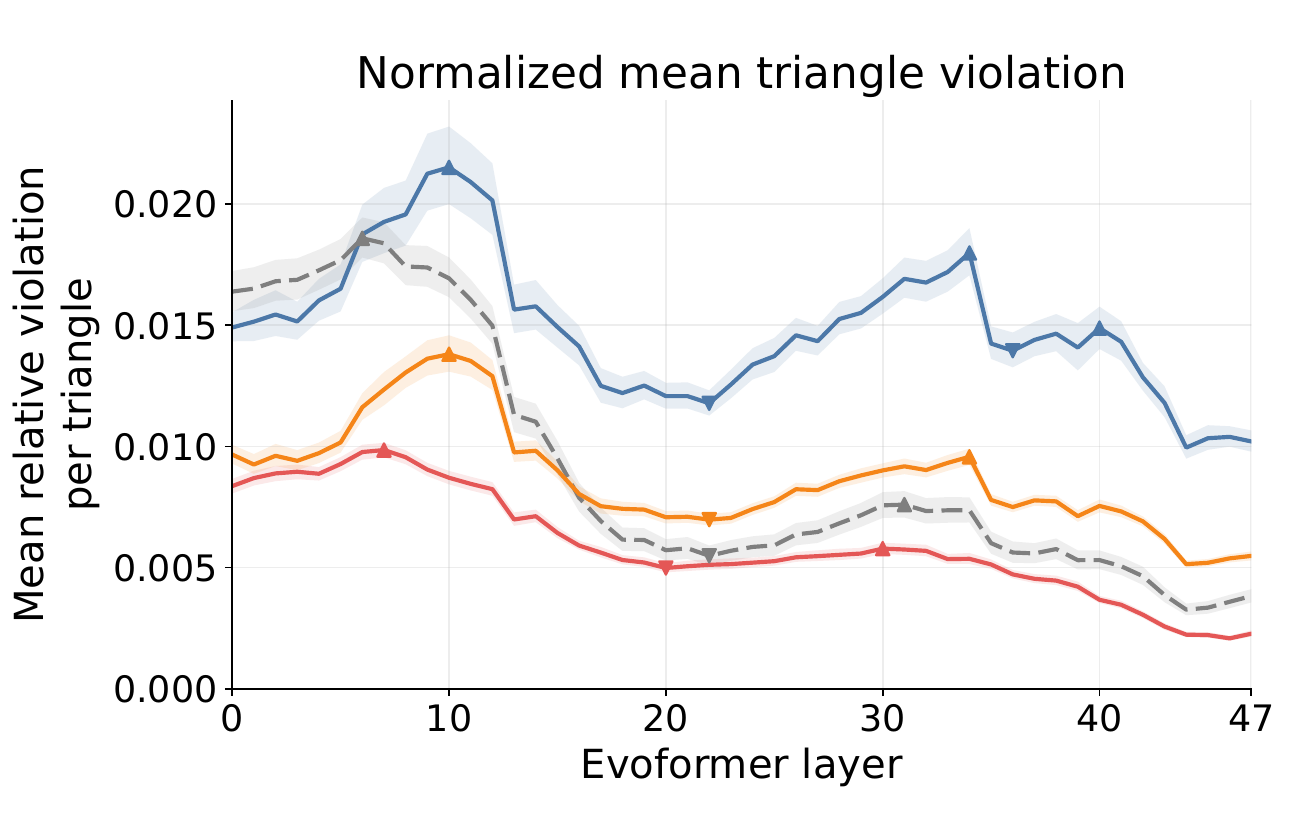}
        \caption{Triangle-violation peaks and troughs}
        \label{fig:triangles-peak-troughs}
    \end{subfigure}\hfill
    \begin{subfigure}[t]{0.49\linewidth}
        \centering
        \includegraphics[width=\linewidth]{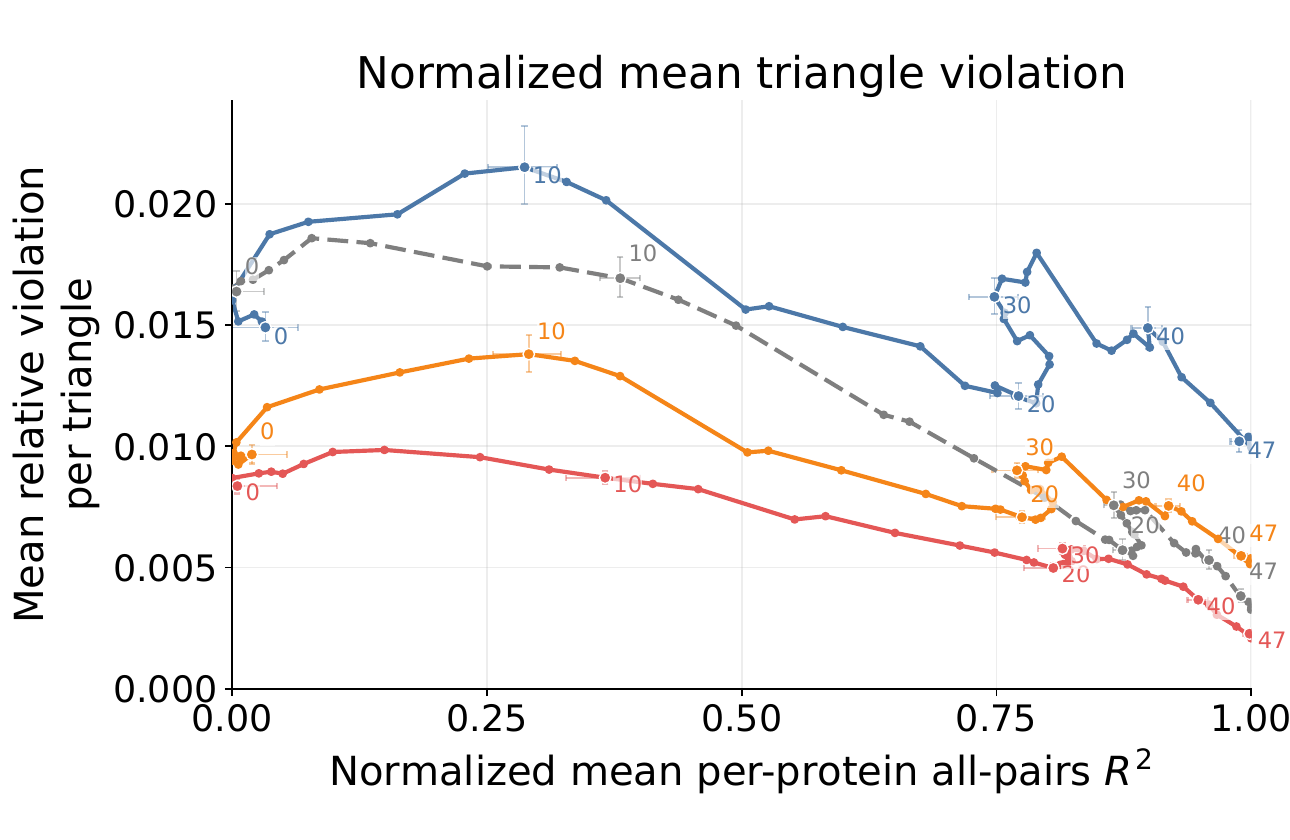}
        \caption{Triangle violations versus decodability}
        \label{fig:triangles-decodability-consistency}
    \end{subfigure}
    \par\smallskip
    \includegraphics[width=\linewidth]{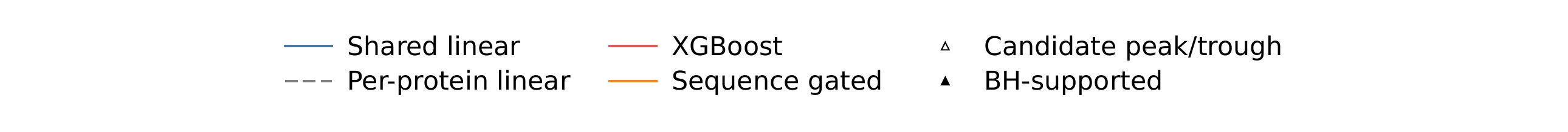}
    \caption{Layerwise triangle-inequality violations and their relationship to distance decodability across the Evoformer. Filled markers denote candidates that remained significant in paired t-tests after Benjamini--Hochberg correction. The x-axis in (b) shows decodability min-max normalized across layers separately for each probe.}
    \label{fig:triangle-violations-and-decodability}
\end{figure}

For each unordered residue triple with decoded sides $\hat a\leq\hat b\leq\hat c$, we defined the triangle-inequality excess as $\delta=\max(\hat c-\hat a-\hat b,\,0)$. Normalized mean violation, our primary measure of triangle violations, is the mean of $\delta$ over all triples, each divided by the RMS side length of the corresponding triangle in the reference structure; Supplement~\S\ref{sec:triangle-defs}. Across all four probes, the normalized mean violation followed the same trajectory (Figure~\ref{fig:triangles-peak-troughs}): a first peak at layers 6--10, a trough at layers 20--22, and a second peak at layers 30--34, followed by a late decline. The pattern was much less pronounced for XGBoost, particularly between layers 20 to 30, where triangle violations rose only modestly. The shared linear probe showed additional supported extrema at layers 36 and 40. All 14 shape-qualified candidate extrema survived Benjamini--Hochberg correction \cite{benjamini1995} (peak and trough detection is described in Supplement~\S\ref{sec:extrema}). Notably, between layers 25 and 30 violations increased while mean $R^2$ decreased for every probe (Figure~\ref{fig:triangles-decodability-consistency}). The violation fraction followed a similar overall pattern, although only 10 of its 13 candidate extrema survived correction (Figure~\ref{fig:supp-triangle-violation-fraction}).

\begin{figure}[t]
    \centering
    \includegraphics[width=\linewidth]{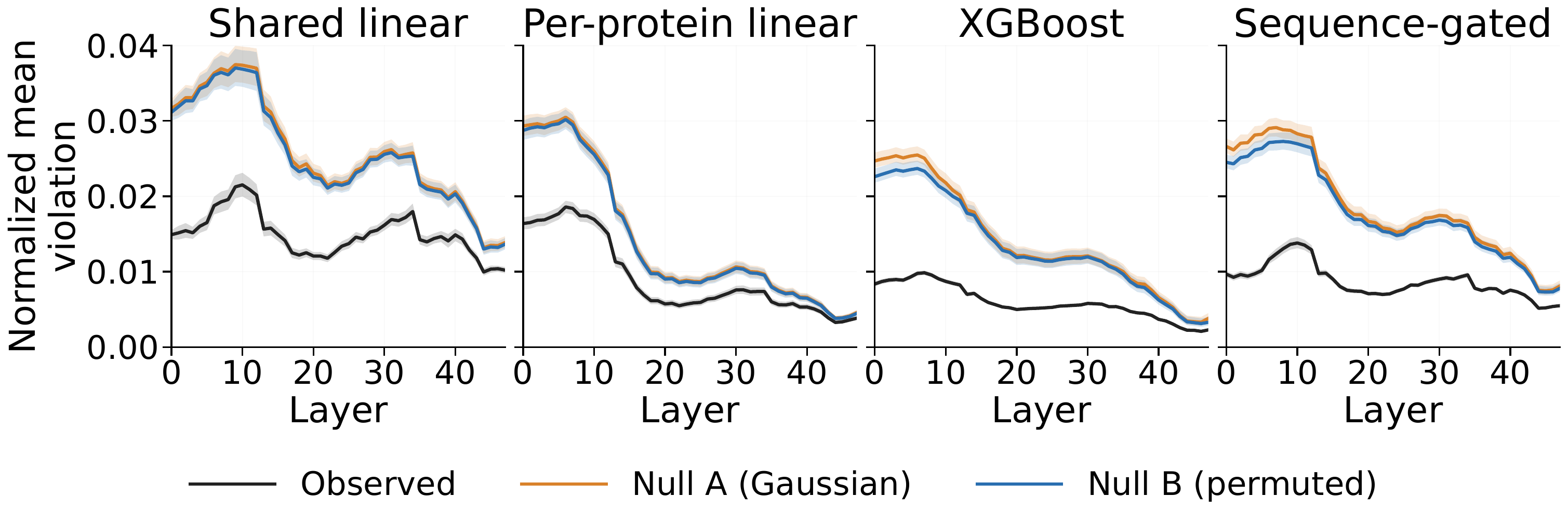}
    \caption{Observed triangle violations compared with two nulls that keep the distribution of probe errors in ten bins of true C$\alpha$--C$\alpha$ distance, but randomize their arrangement. Curves are normalized mean violation, averaged with equal weight across proteins, for all layers and for each probe. Observed decoded distances (black), Null A (Gaussian errors with each spatial bin's mean and standard deviation; orange), and Null B (probe errors permuted within each spatial bin; blue). Shaded bands are 95\% bootstrap confidence intervals over proteins. The nulls are described in Supplement~\S\ref{sec:triangle-nulls}.}
    \label{fig:null-triangles}
\end{figure}

Is the layerwise trend in triangle violations simply a reflection of changes in probe error, or does it also reflect changes in how geometrically coherent the decoded distances are? To explore this, we compared observed violations with two null models matched to distance-binned probe-error statistics. We compute probe errors (predicted minus true C$_\alpha$--C$_\alpha$ distance) for each protein, probe, and layer, and split pairs into 10 spatial distance bins based on true distance. Null A replaces each probe error with an error drawn from a Gaussian distribution with that bin's mean and standard deviation. Null B shuffles the probe errors among pairs within the spatial bin. The resulting errors are added to the true distances and scored with the same triangle metrics. Figure~\ref{fig:null-triangles} (violation fraction: Figure~\ref{fig:supp-null-triangles-fraction}) shows that observed triangle violations are substantially lower than the violations generated from the nulls, at every layer and for every probe. For the shared OLS model, the size of this gap decreases as the bins are made finer, but does not disappear (Figure~\ref{fig:supp-null-triangles-binning}; Supplement~\S\ref{sec:triangle-nulls}). The probe errors are therefore structured rather than random across the edges of a triangle. If independent, randomly distributed probe errors were the source of triangle violations, we would expect the nulls to roughly match observed violations. However, we see a large gap between the nulls and observed values. This implies that the triangle violations we observe are not merely a consequence of the binned error distributions. When combined with the high accuracy of predicted distances and reconstructed structures, particularly in later layers, this supports using decoded distances to measure triangle violations in the pair representation.

The layerwise trends in triangle violations show similarities between the trained probes and nulls, which randomly redistribute the same errors within each distance bin. This makes it unclear if the trend reflects deliberate refinement of geometry in the pair representation. To get more conclusive evidence, we suggest more investigations into whether the reductions in violations are caused by specific components. We leave this to future work.

\subsection{Ablations}

\begin{figure}[t]
    \centering
    \includegraphics[width=\linewidth]{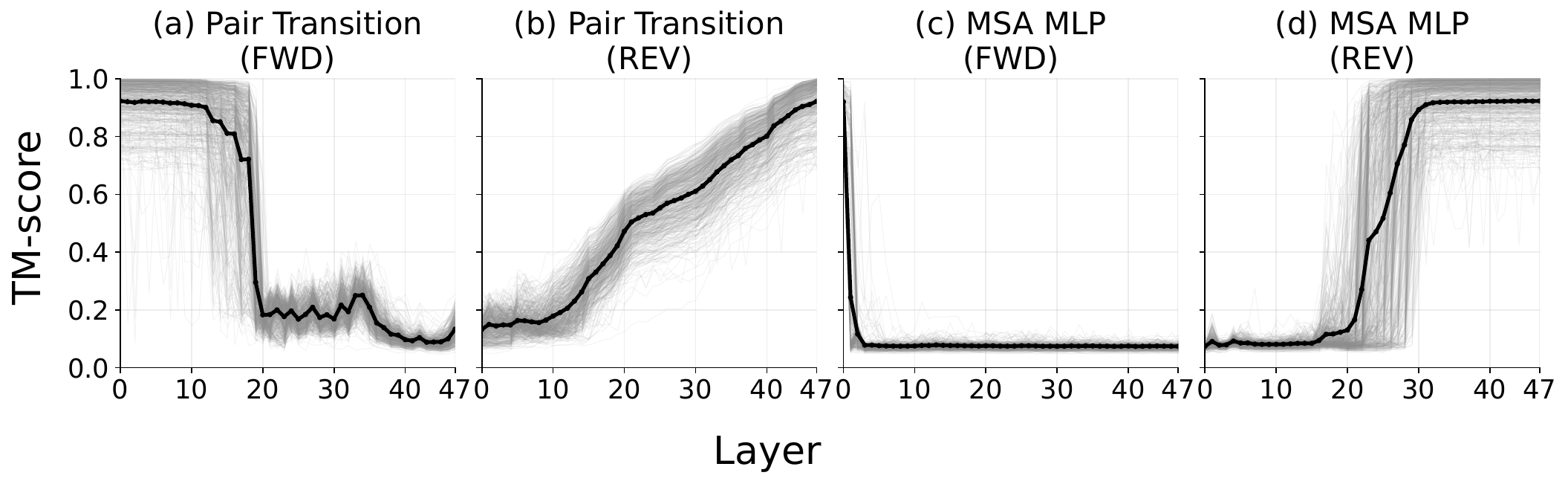}
    \caption{Cumulative layerwise ablations on the combined ECOD+CAMEO cohort. Panels show forward and reverse ablations of the Pair Transition and MSA Transition modules. Light gray traces show individual-protein TM-scores, and the black curve shows the cohort mean.}
    \label{fig:layerwise-ablations-joint}
\end{figure}

To uncover the components of the Evoformer that may be responsible for causing this stage-wise refinement of pairwise distance and triangle consistency, we performed cumulative layer-wise ablations of the Pair Transition MLP, the MSA Transition MLP, and the triangle mechanisms (Triangle Attention and the Triangle Multiplicative Update, bypassed together) by skipping these components for a set of layers in the forward and reverse direction. We then computed the TM-score of each ablated run's final predicted structure against the experimental structure to evaluate the impact of each ablation on structure prediction performance (Figure~\ref{fig:layerwise-ablations-joint}; triangle mechanisms in Figure~\ref{fig:supp-triangle-mechanism-ablations}). Details of the ablations are given in Supplement~\S\ref{sec:ablations}.

Forward ablations, which progressively skip the selected component in layers starting from the beginning, allow us to identify components that are particularly important in early layers of the Evoformer, while reverse ablations, which progressively skip components from the end, reveal which components are crucial in later stages of refining. Forward ablations show that from layers 10 to 20, bypassing the Pair Transition MLP results in a large drop in TM-score of around 0.7 (Figure~\ref{fig:layerwise-ablations-joint}a). This interval where the Pair Transition performs critical computations coincides with a drop in triangle inequality violations. Similarly, forward ablations show that at around layer 30, there is a sudden drop in TM-score when removing triangle mechanisms (Figure~\ref{fig:supp-triangle-mechanism-ablations}a). This corresponds with the reduction in triangle violations we see from our probes. Reverse ablations show a large drop in TM-score when the MSA Transition MLP is removed from the middle layers (Figure~\ref{fig:layerwise-ablations-joint}d). This corresponds to the plateau we see in $R^2$ of decoded distances and indicates that during this phase, evolutionary information may get added or the single representation may get refined in ways that don't directly improve pairwise distance decodability of the pair representation.

\section{Discussion}

Our probes show that Evoformer performs stage-wise processing with two distinct episodes of a decrease in triangle inequality violations. Based on the ablation results, we hypothesize that the Pair Transition MLP and triangle mechanisms are responsible in two distinct phases for improving pairwise distance accuracy and reconciling distance information to increasingly satisfy triangle inequalities. We observe a repeated pattern: triangle violations rise during the same layers where ablating a specific MSA-side component results in a steep drop in TM-score. We postulate that this repeated ``match and correct'' computational pattern involving MSA computations for accessing and updating evolutionary information followed by geometric refinement using triangle mechanisms could be a key step for protein structure determination. A recent work by Lu et al.~\cite{lu2026} also reported a two-stage computational structure across ESMFold, OpenFold, and Boltz-1: early blocks propagate sequence-derived biochemical signals into the pair representation, and late blocks accumulate spatial (distance and contact) information. Our results suggest that geometric refinement is not itself uniform: it contains (at least) two separate episodes of improving geometric consistency, separated by an interval in which decoded triangle consistency worsens and ablating the MSA Transition MLP in successive layers substantially worsens structure prediction accuracy.

Our analysis has several limitations. First, our probes predict the distance using only that pair's own embeddings, $Z_{ij}$ and $Z_{ji}$. Structural information may be distributed across other pair representations, so our analysis measures the information the pair embedding contains for a single pair rather than the total structural information in the representation. Nevertheless, this choice is motivated by the AlphaFold2 architecture and the distogram head, in which pair representations are encouraged to encode pairwise distances and triangle updates are explicitly intended to refine relationships between residue pairs.
Second, the observed trends depend on the capacity of the probes. More expressive models could potentially recover information that our probes do not detect. Conversely, a layer may transform or redistribute structural information without making it more directly decodable as pairwise distance. Thus, our results measure decodability rather than the complete structural information of each layer.
Third, our probes were trained on a single structure prediction model, OpenFold, with recycling disabled, and we only used proteins for which OpenFold's predicted structure had a TM-score $>$ 0.7. Future work exploring other architectures, like Boltz-2 or ESMFold, can help us see whether these trends are replicated in different models.

\clearpage
\setcounter{section}{0}
\setcounter{figure}{0}
\setcounter{table}{0}
\setcounter{equation}{0}
\renewcommand{\thesection}{S\arabic{section}}
\renewcommand{\thefigure}{S\arabic{figure}}
\renewcommand{\thetable}{S\arabic{table}}
\begin{center}{\LARGE\bfseries Supplemental Information}\end{center}
\vspace{1em}
\input{supplemental_information_draft_5_body}

\end{document}

%% file: supplemental_information_draft_5_body.tex
\paragraph{Responsible-use statement.}
Our work analyzes the inner workings of OpenFold, an AI model that predicts protein structures. We release analysis and code only and introduce no new predictive capability, model weights or data. The intended impact of our work is scientific. We aim to offer a clearer picture of how protein structure prediction models work in order to better understand their architecture and the biological rules that determine protein folding. We do not foresee a direct path from these findings to misuse.

\section{Overview}
\label{sec:supp-overview}

In this document, we describe the full methodology for probing the pair representation, reconstructing intermediate 3D structures, and examining triangle inequality consistencies. We also provide additional results that the main paper references and give a more exhaustive list of the analyses we conducted.

\paragraph{Code availability.} Code and instructions for reproducing the analyses are available at \url{https://github.com/ishankhire/match-and-correct-evoformer}.

All experiments and the core methodologies were designed, validated, and tested by the authors. We used LLMs for assistance with writing the scripts and for editing the paper.

\section{Methods}
\label{sec:supp-methods}

\subsection{OpenFold inference and pair representation extraction}
\label{sec:inference}

All representations and predictions were generated with OpenFold
\cite{ahdritz2024}, an open-source implementation of AlphaFold2
\cite{jumper2021} with a 48-block Evoformer trunk. We ran a single forward pass through the 48 layers with recycling disabled. After every layer, we stored the pair representation
\[
Z^{(\ell)}\in\mathbb{R}^{R\times R\times128},
\qquad \ell\in\{0,\ldots,47\},
\]
where $R$ is the residue count.

\subsection{Cohort}
\label{sec:cohort}

To create the analysis cohort, we drew 154 proteins from CAMEO and 400 from ECOD, kept only proteins whose OpenFold prediction scored TM-score $>0.7$ against the experimental structure so that the cohort contains accurate predictions, and then pooled them into a single cohort of 554. Two proteins had incompatible residue counts between
the cached activations and the predicted structure and were dropped, leaving
552 usable proteins. The same 552 proteins are used at every layer and in every analysis reported here, including the ablations of \S\ref{sec:ablations}. The CAMEO and ECOD identifiers of the cohort are listed in \S\ref{sec:protein-list}.

\subsection{Cross-validation designs}
\label{sec:cv}

The 552 eligible proteins were divided into five fixed, approximately equal,
length-stratified folds. For each layer and each shared model, four folds were
used for training and the remaining fold was held out for evaluation, repeated
five times so that every protein is tested exactly once. All preprocessing,
baselines and model fitting were derived only from the four training folds in
each repetition.

Reported performance metrics use only the held-out predictions from each fold.
Training-fold predictions are retained as diagnostics and are not used in any
reported metric. Overall metrics are computed by concatenating the five
disjoint test-fold prediction sets, so every protein, and every evaluated pair,
contributes exactly once. Fold-level means and standard deviations are reported
descriptively. No model is refitted on all proteins after cross-validation.

For the per-protein linear probes, the analogous five-fold procedure is applied
to residue pairs \emph{within} each protein: four pair folds are used for
fitting and the fifth is held out, rotating so that every pair receives an
out-of-fold prediction.

XGBoost hyperparameters were selected before the fivefold analysis using a 60/20/20 fit/validation/test protein split. We compared six candidate configurations that varied maximum tree depth, learning rate, and minimum child weight. Candidates were fit on the fit-split proteins at three anchor layers (0, 23, 47), representing early, middle, and late stages of the Evoformer. For each candidate, we first averaged validation \(R^2\) across proteins at each anchor layer. We then averaged the three layer-level scores and selected the candidate with the highest value. Candidate configurations performed very similarly, so the
precise winner was not a strong qualitative choice. That fixed configuration (\S\ref{sec:probes}) was then trained independently within each of the five folds. XGBoost used up to
5{,}000 sampled training pairs per fitting protein but evaluated all pairs in
each held-out protein.

\subsection{Regression features and targets}
\label{sec:features}

Each regression sample is a residue pair drawn from the Evoformer pair representation. The target is
\[
y_{ij} = \| c_i - c_j \|_2,
\]
where $c_i \in \mathbb{R}^3$ is the final predicted $C_\alpha$ coordinate of
residue $i$ in the structure produced by OpenFold's Structure Module. The
target is therefore the model's own output, not an experimentally derived
structure, and is fixed across layers while only the input feature vector
changes with $\ell$.

The base feature for layer $\ell$ concatenates both directional entries:
\[
x_{ij}^{(\ell)} =
\left[Z_{ij}^{(\ell)} \;\|\; Z_{ji}^{(\ell)}\right]
\in \mathbb{R}^{256}, \qquad i > j.
\]

\subsection{Preprocessing and objective}
\label{sec:fitting}

Before fitting each OLS probe, we standardized features using means and standard deviations estimated exclusively from that model's training observations. Shared OLS models used protein-equal weighted statistics, sequence-separation-gated models used separate protein-equal statistics within each bin, and per-protein models used statistics from that protein's training pairs. Each fitted transformation was applied in the same way to all non-training samples. XGBoost features were not standardized.

For shared models, each protein received equal total weight during fitting. Because the number of residue pairs grows quadratically with protein length, we averaged the squared error across pairs within each protein and then averaged these losses equally across proteins to prevent longer ones from dominating. The resulting objective was

\[
\mathcal{L} =
\frac{1}{P} \sum_{p=1}^{P}
\frac{1}{n_p} \sum_{(i,j) \in p}
\left(y_{pij} - \hat y_{pij}\right)^2.
\]

where \(P\) is the protein count and \(n_p\) is the number of selected training pairs of protein \(p\).

\subsection{Probing models}
\label{sec:probes}

\paragraph{Shared linear probe.} At each Evoformer layer, an ordinary
least-squares model with intercept is fitted on the training-fold proteins,
predicting final-structure $C_\alpha$--$C_\alpha$ distance from the
256-dimensional concatenated directional pair features.

\paragraph{Sequence-separation-gated probe.} This model followed the same fitting procedure but assigned residue pairs into six predefined bins based on their separation $|i-j|$ along the protein sequence: 1, 2--4, 5--9, 10--19, 20--49, and 50 or more residues. A separate shared linear probe is fitted within each bin and the six predictions
are recombined before whole-protein metrics are computed.

\paragraph{Per-protein linear probe.} A separate linear model is fitted for each
protein at each layer, using the within-protein pair folds described in
\S\ref{sec:cv}. This measures generalisation to unseen residue pairs within a
known protein rather than transfer to unseen proteins.

\paragraph{Nonlinear probe (XGBoost).} A gradient-boosted tree regressor
with a squared-error objective \cite{chen2016} is fitted at each layer on the same 256-dimensional features
with protein-equal weighting. The selected configuration (\S\ref{sec:cv}) uses maximum tree
depth 6, learning rate 0.1, minimum child weight 10, row and feature
subsampling rates 0.8, an $L_2$ penalty of 1.0, a deterministic cap of 5{,}000
sampled residue pairs per fitting protein, and a fixed budget of 700 boosting
rounds. Evaluation uses all available pairs.

\subsection{Accuracy metrics}
\label{sec:metrics}

\textbf{Mean per-protein $R^2$} is the fraction of distance variance explained
within a protein, averaged equally across proteins. This is the headline
measure because it gives each protein equal weight. \textbf{Pooled $R^2$} pools
all residue pairs before computing variance explained, so larger proteins have
more influence. \textbf{MAE} and \textbf{RMSE} are reported in \aa ngstroms.

$R^2$ becomes unstable when the target barely varies, which is the regime for
adjacent residues: $C_\alpha$--$C_\alpha$ distance at sequence separation 1
averages almost exactly 3.8\,\AA{} with a very low standard deviation. MAE is the interpretable metric there, which is why the
stratified analyses (main-text Figure~1c and \S\ref{sec:strat-acc}) report MAE rather than $R^2$.

\paragraph{Stratified accuracy.} Stratified accuracy evaluates the same held-out predictions used for the whole-protein metrics; no category-specific models are fitted. Within each protein and category, MAE is the mean of $|y_{ij}-\hat y_{ij}|$ over the pairs in that category, and curves average these per-protein values with equal protein weight. Sequence-separation bins are the six gates of \S\ref{sec:probes}. Spatial-distance bins use the true $C_\alpha$--$C_\alpha$ distance in OpenFold's final structure, with canonical bins of 0--5, 5--10, 10--20, 20--40, and at least 40~\AA. The 10--40~\AA{} category shown in main-text Figure~1c combines the 10--20 and 20--40~\AA{} bins by summing each protein's absolute-error totals and pair counts before dividing. True distance is used only to stratify evaluation, never to route training or prediction.

\paragraph{Bootstrap intervals.} Uncertainty on every protein-equal mean curve is estimated by a paired-protein bootstrap. The 552 proteins are resampled with replacement 2{,}000 times with a fixed seed, the protein-equal mean is recomputed in each replicate at every layer, and the pointwise 2.5th and 97.5th percentiles of the replicate means give the 95\% interval. Because whole proteins are resampled, the same replicate is used at every layer, so the bands are paired across layers. In the decodability--consistency phase plots (main-text Figure~2b and Figure~\ref{fig:supp-triangle-fraction-decodability}), each protein's $R^2$ and triangle statistics are resampled jointly, and horizontal and vertical bars are drawn at layers 0, 10, 20, 30, 40, and 47. These intervals describe sampling variability across proteins in the cohort and are not the fold-level standard deviations of \S\ref{sec:cv}.

\subsection{MDS reconstruction}
\label{sec:mds}

We reconstructed 3D C$\alpha$ coordinates from the predicted pairwise distances using classical multidimensional scaling (MDS) \cite{torgerson1952}. Negative predicted pairwise distances from the linear model were clipped to 0 before performing MDS. For each protein, the squared predicted distance matrix was double-centered to obtain an estimated Gram matrix,
\[
B = -\frac{1}{2} J D^{(2)} J,
\]
where \(D^{(2)}\) denotes the elementwise squared distance matrix and \(J\) is the centering matrix. The reconstructed C\(\alpha\) coordinates were then given by
\[
X_3 = Q_3 \Lambda_3^{1/2},
\]
where \(Q_3\) contains the eigenvectors associated with the three largest positive eigenvalues of \(B\), and \(\Lambda_3\) is the corresponding diagonal eigenvalue matrix. Thus, MDS provides a rank-3 Euclidean approximation to the distance geometry predicted by the probe.

Because classical MDS recovers coordinates only up to a rigid motion and a reflection, we evaluated both the original and the mirror-image three-dimensional solutions against the C$\alpha$ coordinates of OpenFold's final predicted structure. For RMSD, we used the Kabsch algorithm and retained the lower-RMSD alignment. Independently, we optimized the TM-score \cite{zhangskolnick2004} with the standard TMscore executable, keeping residue correspondence fixed, normalizing by the reference protein length, and retaining the highest score across both reflections and their Kabsch candidates. The executable's rigid transformations were evaluated on the original double-precision coordinates; this heuristic search does not guarantee a global maximum.

\begin{figure}[t]
  \centering
  \begin{subfigure}[t]{0.49\linewidth}
    \centering
    \includegraphics[width=\linewidth]{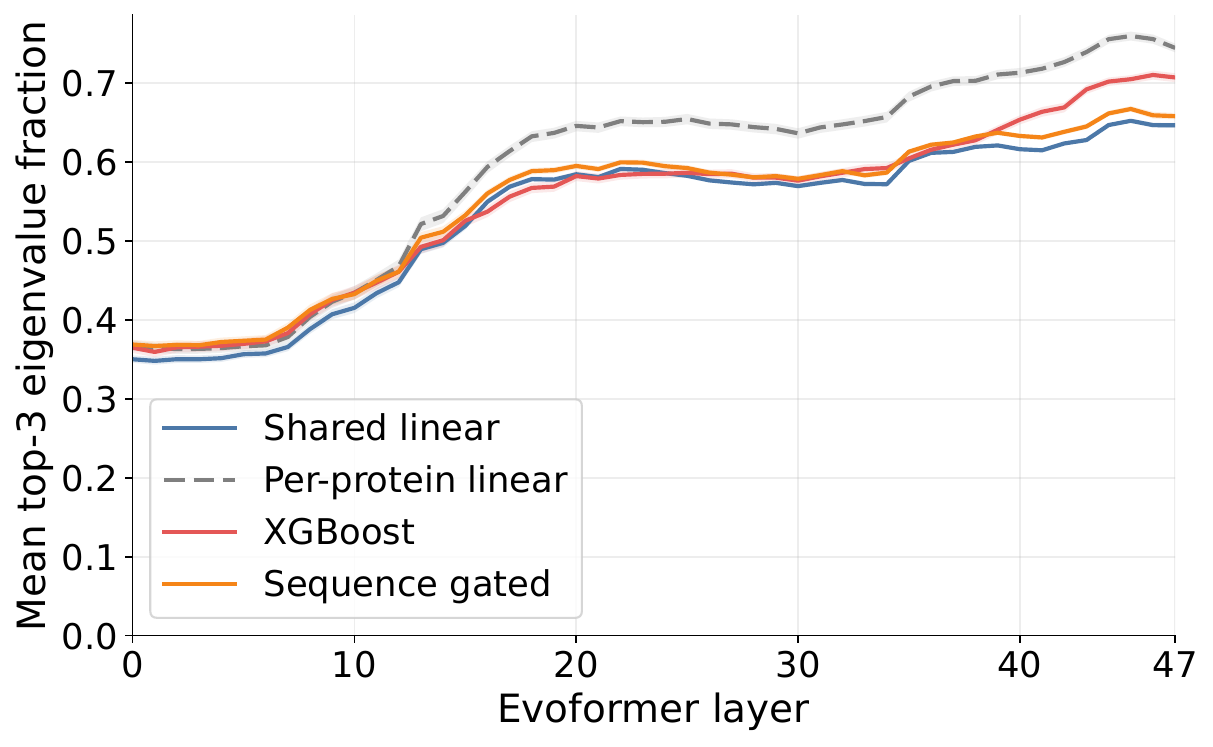}
    \caption{Top-three positive-eigenvalue fraction}
    \label{fig:supp-top3-eigenvalue-fraction}
  \end{subfigure}\hfill
  \begin{subfigure}[t]{0.49\linewidth}
    \centering
    \includegraphics[width=\linewidth]{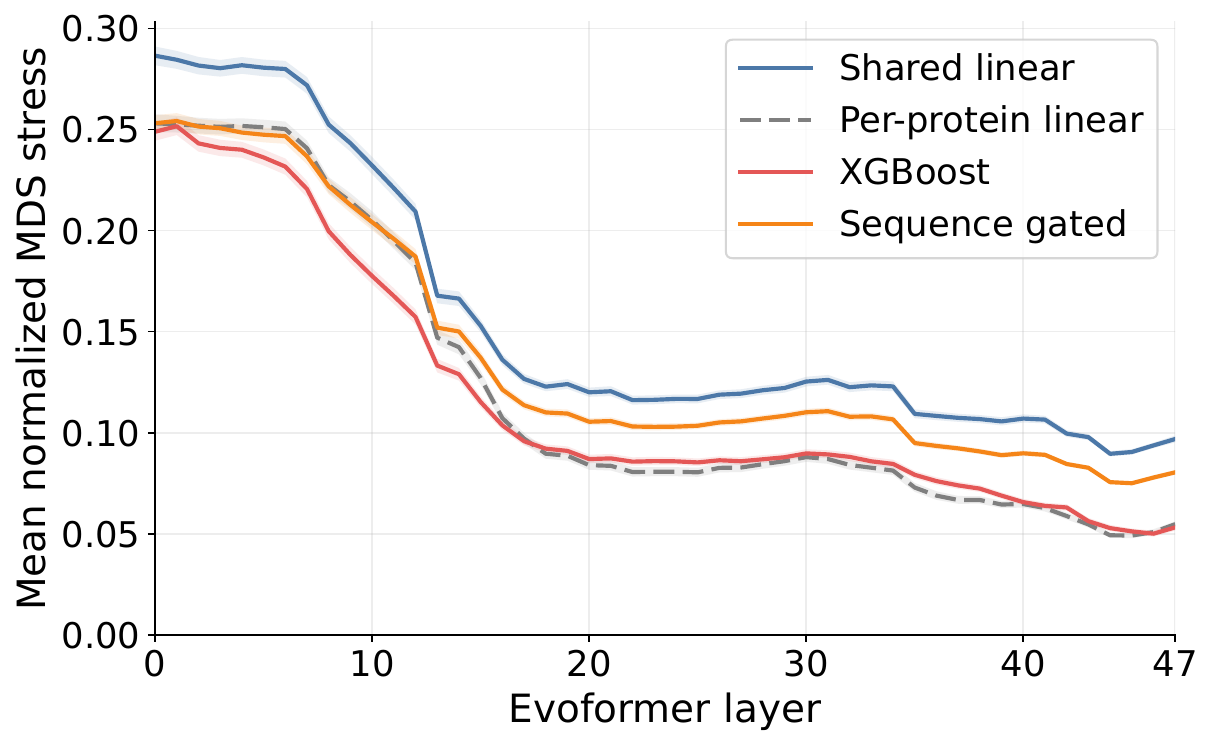}
    \caption{Normalized MDS stress}
    \label{fig:supp-mds-stress}
  \end{subfigure}
  \caption{Classical-MDS geometry diagnostics across Evoformer layers. (a) The top-three positive-eigenvalue fraction is $\sum_{q=1}^{3}\max(\lambda_q,0)/\sum_{\lambda_q>0}\lambda_q$ for the eigenvalues of the double-centered squared predicted distance matrix, ordered from largest to smallest. (b) Normalized MDS stress is $\sqrt{\sum_{i<j}(\hat d_{ij}-d^{\mathrm{MDS}}_{ij})^2/\sum_{i<j}\hat d_{ij}^{2}}$, where $\hat d_{ij}$ is the probe-predicted distance after clipping negative predictions to zero and $d^{\mathrm{MDS}}_{ij}$ is the distance in the three-dimensional reconstruction. Curves are equal-weight means over 552 proteins and bands are paired-protein bootstrap percentile 95\% intervals from 2,000 replicates.}
  \label{fig:mds-geometry-diagnostics}
\end{figure}

Because MDS always returns a point cloud after discarding incompatible spectral
components, a good reconstruction does not establish that the decoded matrix
was a valid Euclidean distance matrix. We therefore report two diagnostics
per protein and layer, shown in Figure~\ref{fig:mds-geometry-diagnostics}:

\begin{enumerate}[leftmargin=*,itemsep=1pt]
\item \textbf{Top-three positive eigenvalue mass}: the fraction of the sum of
positive eigenvalues of $B$ carried by the three largest, which is 1 for an
exactly rank-3 configuration.
\item \textbf{Three-dimensional reconstruction stress}: the RMS discrepancy
between decoded distances and the distances realised in the rank-3
reconstruction, divided by the RMS decoded distance.
\end{enumerate}

We display a subset of the predicted structures we were able to generate using MDS and their progression across layers. Studying how these structures evolve may assist future research projects in uncovering the workings of the Evoformer (Figure~\ref{fig:mds-structures}).

\subsection{Triangle-inequality analysis}
\label{sec:triangle-defs}

For each protein, layer, and probe, we assembled the predicted distances into a symmetric \(R \times R\) matrix with zero diagonal. Negative predictions were recorded and then clipped to zero. Every unordered residue triple is enumerated exactly once, giving $\binom{R}{3}$ triples for a protein of $R$ residues. For each triple the three decoded sides are sorted as $\hat a \le \hat b \le \hat c$, and the violation magnitude is
\begin{equation}
\delta = \max(\hat c - \hat a - \hat b,\, 0),
\end{equation}
with a triple counted as violating only when $\hat c - \hat a - \hat b > 10^{-12}$~\AA.

For a category containing $N$ triples we report
\begin{align}
\text{violation fraction} &= \frac{\#\{\delta > 0\}}{N}, \\[4pt]
\text{normalized mean violation} &= \frac{1}{N}\sum_{t=1}^{N} \frac{\delta_t}{r_t},
\end{align}
where $r_t$ is the RMS side length of triple $t$ in the \emph{reference} structure (\S\ref{sec:strata}), so non-violating triples contribute zero to the second mean. The normalizing denominator is taken from the reference rather than the decoded triangle so that it is fixed across layers and probes. We additionally record severity fractions: the fractions of all triples with $0 < \delta < 5$, $5 \le \delta < 10$, and $\delta \ge 10$~\AA, which sum to the violation fraction. All quantities are computed per protein and then averaged equally across proteins.

\subsection{Reference triangle categories}
\label{sec:strata}

All categories are fixed from each protein's C$\alpha$ geometry in OpenFold's final predicted structure. They are not recomputed from decoded distances and do not vary by probe or layer, so a category contains the same triples in every panel of every figure. For each unordered triple, let $x_i, x_j, x_k \in \mathbb{R}^3$ be the reference coordinates and let the three reference side lengths be sorted as $a \le b \le c$, so the largest interior angle is opposite $c$.

\begin{itemize}[leftmargin=*,itemsep=1pt]
\item \textbf{All triangles}: every unordered residue triple.
\item \textbf{RMS size}: $r = \sqrt{(a^2+b^2+c^2)/3}$, in \aa ngstroms. Small, medium, and large are the within-protein tertiles of $r$.
\item \textbf{Area}: $A = \tfrac{1}{2}\|(x_j - x_i) \times (x_k - x_i)\|$. Low, medium, and high are the within-protein tertiles of $A$.
\item \textbf{Relative slack}: $s = (a + b - c)/c \in [0,1]$. Low, medium, and high are the within-protein tertiles of $s$.
\item \textbf{Largest angle}: $\theta = \cos^{-1}\!\big((a^2+b^2-c^2)/(2ab)\big)$. Clearly acute is $\theta < 88^\circ$, near-right is $88^\circ \le \theta \le 92^\circ$, and clearly obtuse is $\theta > 92^\circ$.
\end{itemize}

For the three tertile systems, with within-protein quantiles $q_1$ and $q_2$, the half-open bins are $x < q_1$, $q_1 \le x < q_2$, and $x \ge q_2$. Each category system partitions the full set of triples. These five systems give 13 trajectories per probe.

\subsection{Peak and trough detection}
\label{sec:extrema}

Peaks and troughs are tested on the two triangle outcomes of \S\ref{sec:triangle-defs} for each of the 13 trajectories of each probe. $R^2$ and other decodability measures are not tested.

\paragraph{Eligibility and shape.} Only center layers $\ell \in \{5,\dots,42\}$ are eligible, so that five layers are available on each side. A candidate peak at $\ell$ must have a protein-equal mean $\bar y_\ell$ strictly greater than each of the ten neighboring layer means at $\ell-5,\dots,\ell-1,\ell+1,\dots,\ell+5$, and OLS lines fitted to the six means at $\ell-5,\dots,\ell$ and at $\ell,\dots,\ell+5$ must have positive and negative slope, respectively. A candidate trough reverses all inequalities and slope signs. These are shape requirements only, not significance tests.

\paragraph{Paired test.} For each protein $p$, define the center-versus-neighborhood contrasts
\begin{equation}
D_{L,p} = y_{p,\ell} - \frac{1}{5}\sum_{j=\ell-5}^{\ell-1} y_{p,j},
\qquad
D_{R,p} = y_{p,\ell} - \frac{1}{5}\sum_{j=\ell+1}^{\ell+5} y_{p,j}.
\end{equation}
For a peak, separate one-sided one-sample $t$-tests across the 552 proteins (551 degrees of freedom) test $E[D_L] > 0$ and $E[D_R] > 0$; for a trough the same tests are applied to $-D_L$ and $-D_R$. Because a supported extremum must exceed both neighborhoods, the two tests form an intersection--union test with joint $p$-value $p_{\mathrm{joint}} = \max(p_L, p_R)$. Proteins, not triples, are the independent units. Every eligible layer is tested in both directions before the shape filter is applied.

\paragraph{Multiple comparisons.} Benjamini--Hochberg correction \cite{benjamini1995} is applied separately to each triangle outcome, pooling all probes, categories, eligible layers, and directions into one family of
\begin{equation}
4\ \text{probes} \times 13\ \text{trajectories} \times 38\ \text{layers} \times 2\ \text{directions} = 3{,}952
\end{equation}
hypotheses. An extremum is reported as supported only if it is shape-qualified and has $q \le 0.05$. In the figures, hollow upward and downward triangles mark shape-qualified candidate peaks and troughs, and filled markers denote candidates that are also BH-supported.

\subsection{Null models for triangle violations}
\label{sec:triangle-nulls}

To test whether the triangle violations of the decoded distances are explained by the size of the probe errors alone, we compared them with two null models that keep each protein's probe errors but rearrange them among residue pairs. For each protein, probe, and layer, let $e_{ij}=\hat d_{ij}-d_{ij}$ be the error of the decoded distance relative to the reference C$\alpha$--C$\alpha$ distance in OpenFold's final predicted structure. Residue pairs are grouped into ten half-open bins of $d_{ij}$ with edges at 0, 4, 6, 8, 10, 14, 20, 30, 40, and 55~\AA.

\begin{itemize}[leftmargin=*,itemsep=1pt]
\item \textbf{Null A (Gaussian)}: each error is replaced by a draw from a normal distribution with the mean and standard deviation of that protein's errors in the same bin. A bin with fewer than 30 pairs in a protein instead uses the protein-equal mean of these statistics over proteins with at least 30 pairs in the bin.
\item \textbf{Null B (permuted)}: the errors are randomly permuted among the pairs within each bin, so every bin keeps its exact error distribution.
\end{itemize}

The null distances $d_{ij}+\tilde e_{ij}$ are scored with the metrics of \S\ref{sec:triangle-defs}, including clipping negative distances to zero, on every unordered residue triple, exactly as the decoded distances are. Each null was drawn three times per protein, probe, and layer, and the draws were averaged within each protein; the Monte Carlo error of the resulting means is negligible next to the bootstrap intervals. To check that the gap between observed and null violations does not depend on the choice of bins, we repeated Null B for the shared linear probe with 5 bins (edges at 0, 5, 10, 20, and 40~\AA), with 20 bins (each of the ten bins split at its midpoint and the last at 70~\AA), and with the ten distance bins crossed with the six sequence-separation bins of \S\ref{sec:probes}.

\begin{figure}[t]
  \centering
  \includegraphics[width=\linewidth]{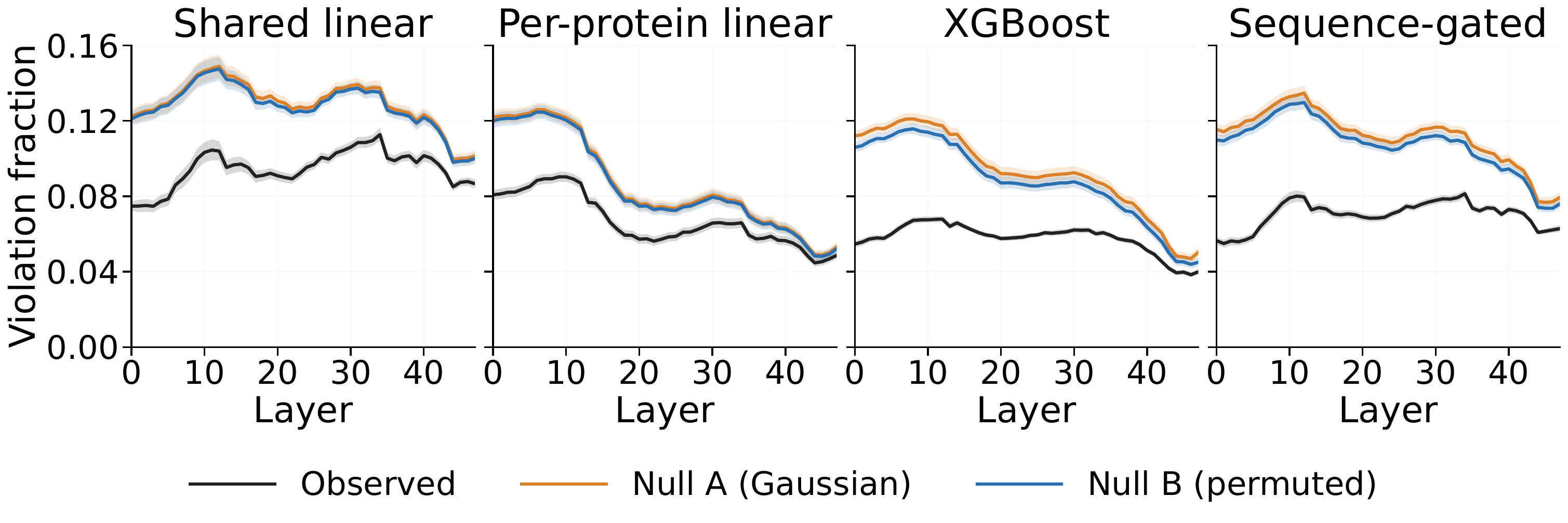}
  \caption{Triangle-inequality violation fraction of the observed decoded distances (black) compared with Null A (orange) and Null B (blue), using ten spatial bins, for every unordered residue triple. Curves are equal-weight means over 552 proteins and bands are paired-protein bootstrap percentile 95\% intervals from 2,000 replicates.}
  \label{fig:supp-null-triangles-fraction}
\end{figure}

\begin{figure}[t]
  \centering
  \includegraphics[width=\linewidth]{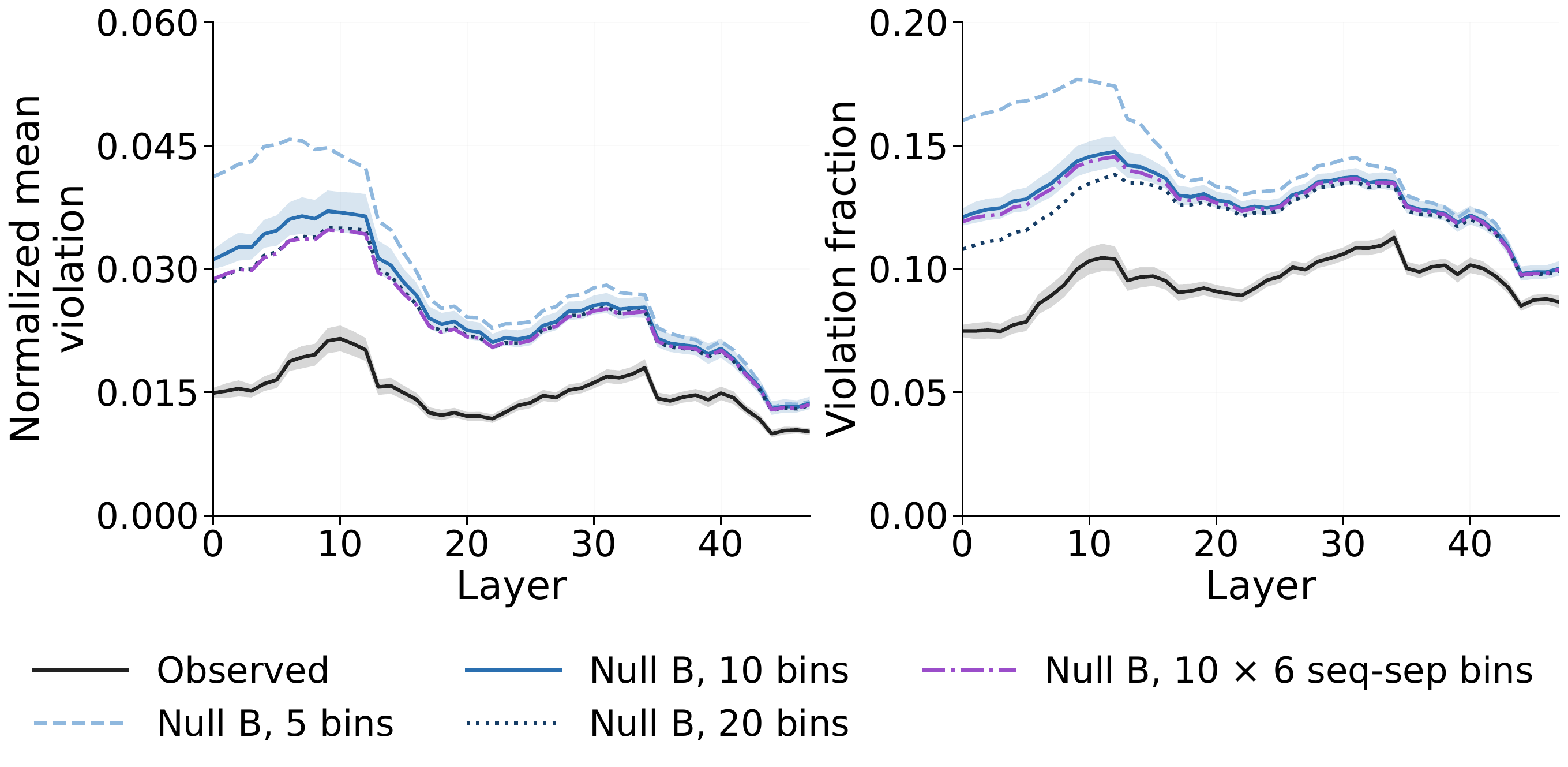}
  \caption{Sensitivity of Null B to the choice of bins, for the shared linear probe. Left: normalized mean violation; right: violation fraction. Observed decoded distances (black) are compared with Null B using 5, 10, or 20 bins of true C$\alpha$--C$\alpha$ distance, or the 10 distance bins crossed with 6 sequence-separation bins. Bands (shown for observed and for Null B with 10 bins) are paired-protein bootstrap percentile 95\% intervals over 552 proteins.}
  \label{fig:supp-null-triangles-binning}
\end{figure}

Figure~\ref{fig:supp-null-triangles-fraction} shows the violation-fraction companion of main-text Figure~\ref{fig:null-triangles}, and Figure~\ref{fig:supp-null-triangles-binning} shows the binning comparison.

\subsection{Cumulative layerwise ablations}
\label{sec:ablations}

The ablations are inference-time component-skipping experiments run on the same zero-recycle OpenFold configuration as \S\ref{sec:inference}. Three component families are considered: the Pair Transition MLP; the MSA Transition MLP (labeled ``MSA MLP'' in figures); and the triangle mechanisms, meaning starting-node Triangle Attention, ending-node Triangle Attention, and the Triangle Multiplicative Update bypassed together. Bypassing a module means that its update is not added to the residual stream in the designated blocks; residual connections, every other Evoformer module, all later blocks, and the Structure Module execute normally.

Ablations are cumulative in one of two directions. In the \emph{forward} series, boundary $k \in \{0,\dots,47\}$ bypasses the component in blocks $0,\dots,k$ (skip the module in layer 0, then in layers 0 and 1, and so on). In the \emph{reverse} series, boundary $k$ bypasses it in blocks $k,\dots,47$ (skip the module in layer 47, then in layers 47 and 46, and so on). A point at boundary $k$ therefore reflects the cumulative intervention up to that boundary, not the isolated effect of block $k$. For each protein, component family, direction, and boundary, one complete OpenFold inference was run; because of the computational cost, each layerwise ablation was run once per protein.

Each ablated run is scored by the TM-score of its final predicted structure against the chain-matched experimental structure from the RCSB, computed with TM-align \cite{zhang2005tmalign}. This target differs from the probe--MDS TM-score of \S\ref{sec:mds}, which compares an MDS reconstruction with OpenFold's own unablated final prediction. Ablation curves are restricted to the 552-protein probe cohort of \S\ref{sec:cohort}; thin gray traces are individual proteins and the black curve is the protein-equal mean.

\section{Supplemental figures}
\label{sec:figures}

\subsection{Triangle violations stratified by reference geometry}

\begin{figure}[t]
  \centering
  \includegraphics[width=\linewidth]{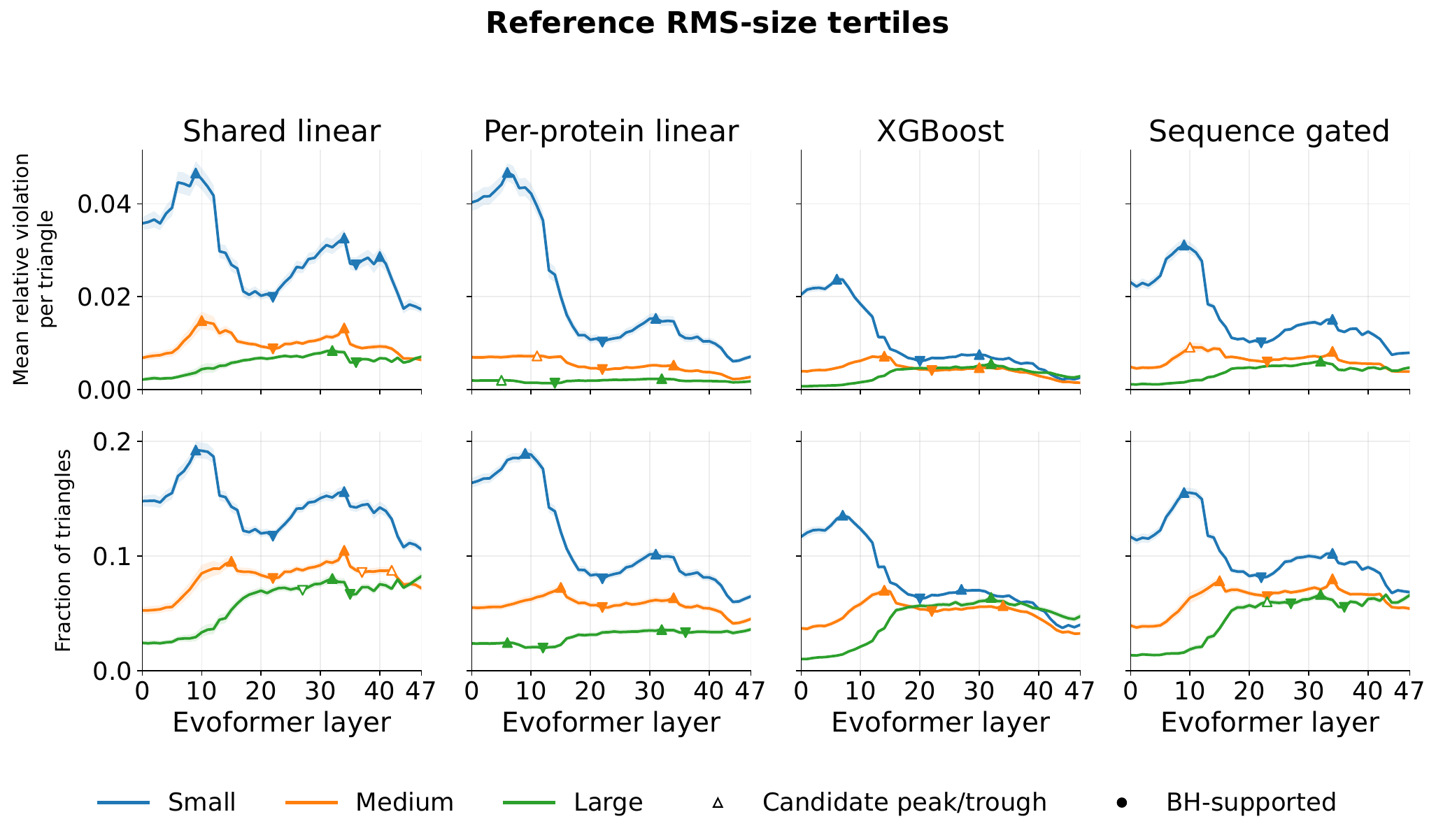}
  \caption{Triangle-inequality metrics stratified by reference RMS-size tertile. RMS size is $\sqrt{(a^2+b^2+c^2)/3}$ for the triangle in OpenFold's final predicted structure; small, medium, and large denote within-protein tertiles of this quantity.}
  \label{fig:supp-triangle-rms-size}
\end{figure}

\begin{figure}[t]
  \centering
  \includegraphics[width=\linewidth]{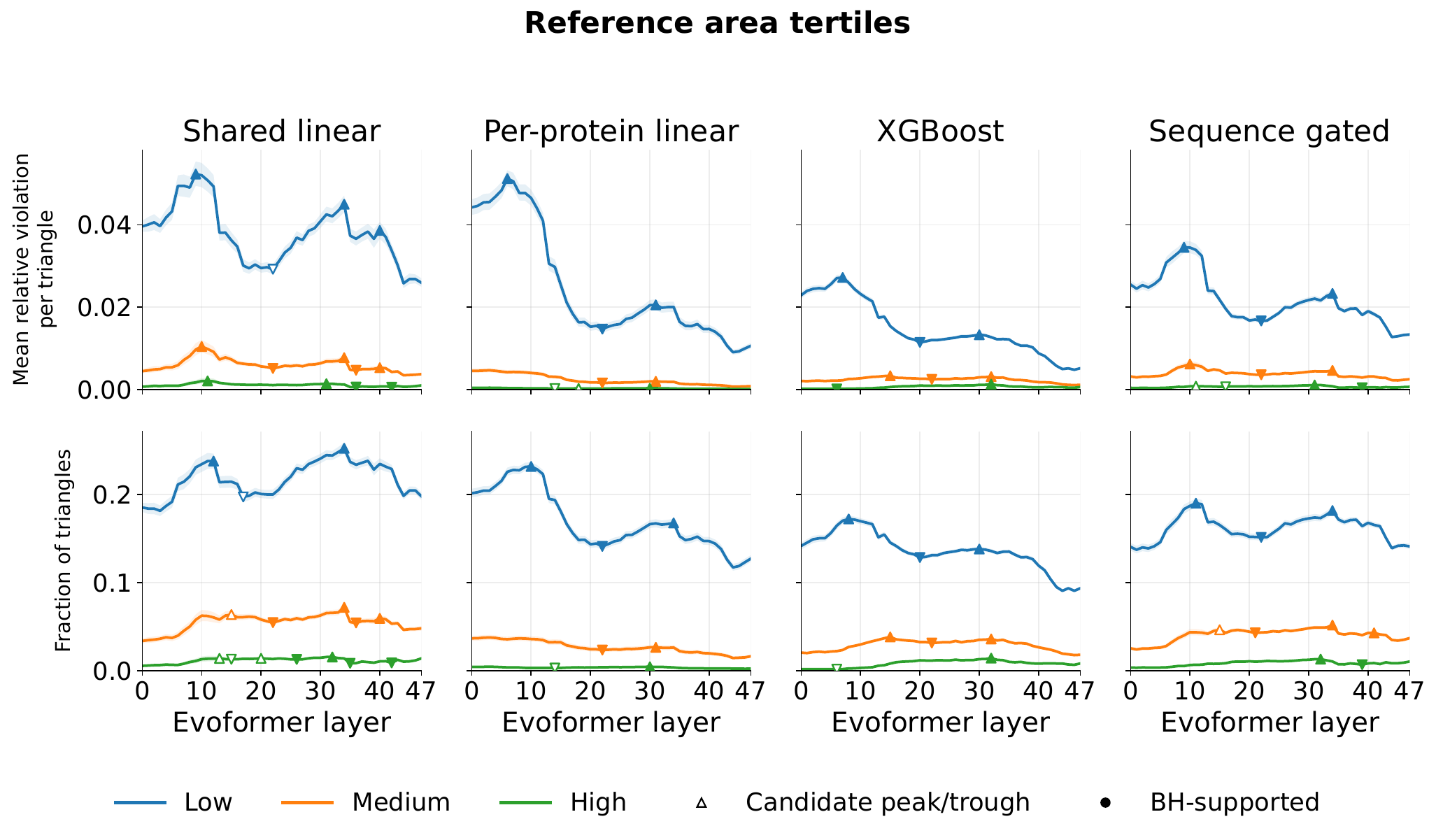}
  \caption{Triangle-inequality metrics stratified by reference-area tertile. Area is $\frac{1}{2}\lVert(x_j-x_i)\times(x_k-x_i)\rVert$ for the C$\alpha$ triangle in OpenFold's final predicted structure; low, medium, and high denote within-protein tertiles.}
  \label{fig:supp-triangle-area}
\end{figure}

\begin{figure}[t]
  \centering
  \includegraphics[width=\linewidth]{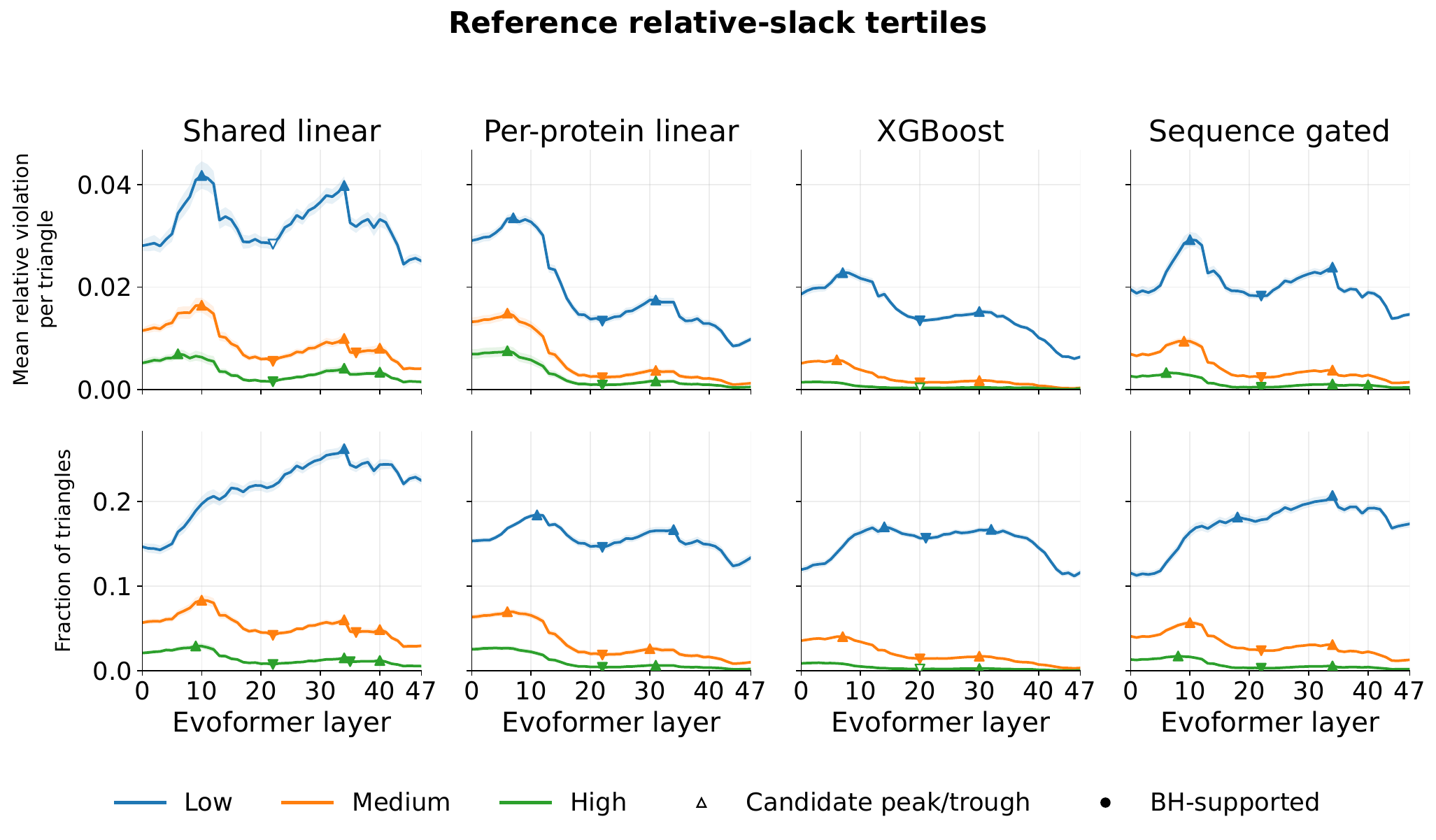}
  \caption{Triangle-inequality metrics stratified by reference relative-slack tertile. Relative slack is $(a+b-c)/c$ for the sorted side lengths $a \le b \le c$ of the triangle in OpenFold's final predicted structure; low, medium, and high denote within-protein tertiles.}
  \label{fig:supp-triangle-relative-slack}
\end{figure}

\begin{figure}[t]
  \centering
  \includegraphics[width=\linewidth]{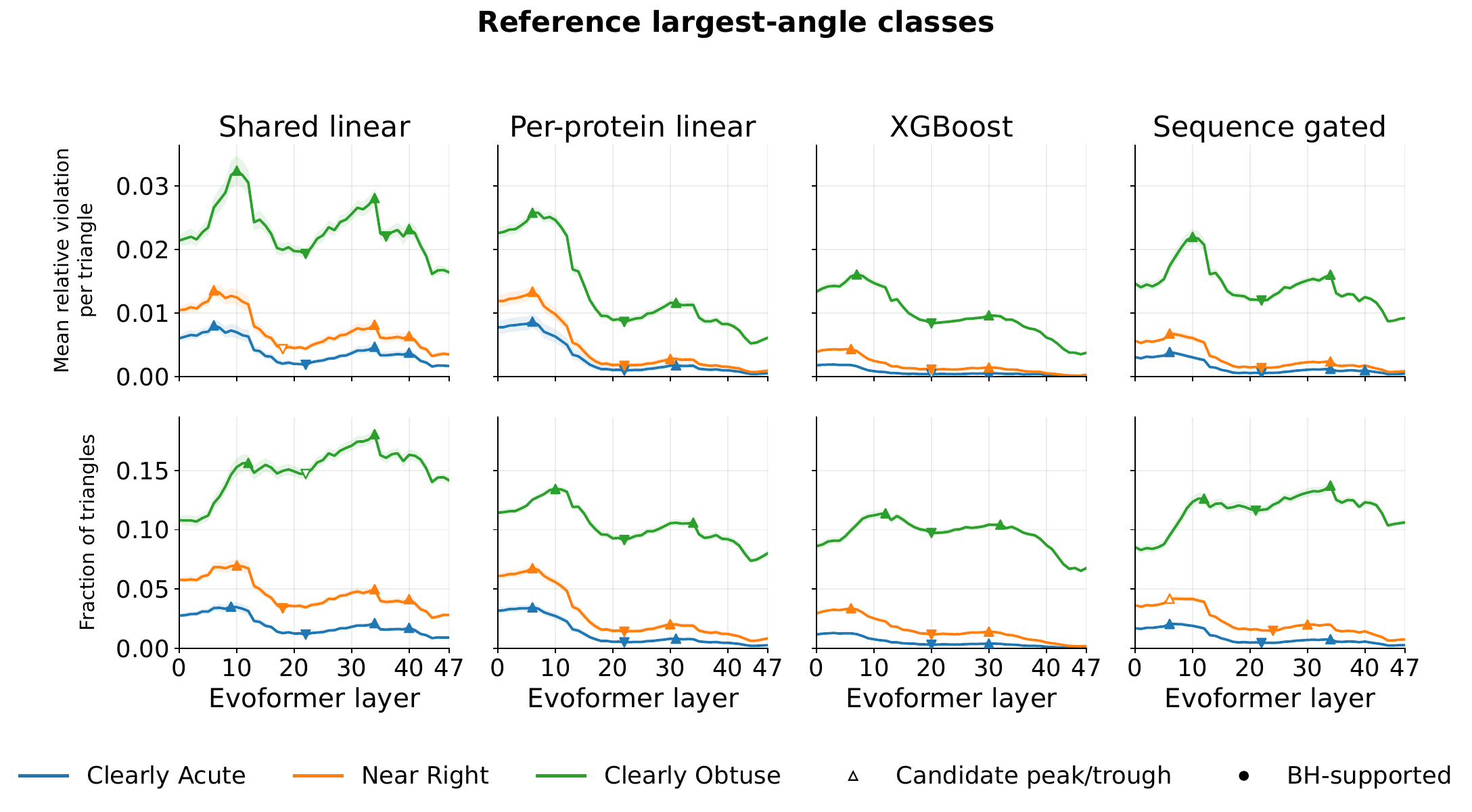}
  \caption{Triangle-inequality metrics stratified by the largest reference angle. The largest angle is opposite $c$ and equals $\arccos\big((a^2+b^2-c^2)/(2ab)\big)$ for the triangle in OpenFold's final predicted structure. Clearly acute denotes angles below $88^\circ$, near-right denotes angles from $88^\circ$ through $92^\circ$, and clearly obtuse denotes angles above $92^\circ$.}
  \label{fig:supp-triangle-angle-shape}
\end{figure}

Figures~\ref{fig:supp-triangle-rms-size}--\ref{fig:supp-triangle-angle-shape} show the two triangle outcomes of \S\ref{sec:triangle-defs} within the four stratification systems of \S\ref{sec:strata}. In each figure, columns are the four probe families, the top row is normalized mean violation, and the bottom row is violation fraction. Curves are equal-weight means over 552 proteins, bands are paired-protein bootstrap percentile 95\% intervals from 2,000 replicates, and peak and trough markers follow \S\ref{sec:extrema}.

\clearpage
\subsection{Violation fraction for all triangles}

\begin{figure}[t]
  \centering
  \includegraphics[width=0.86\linewidth]{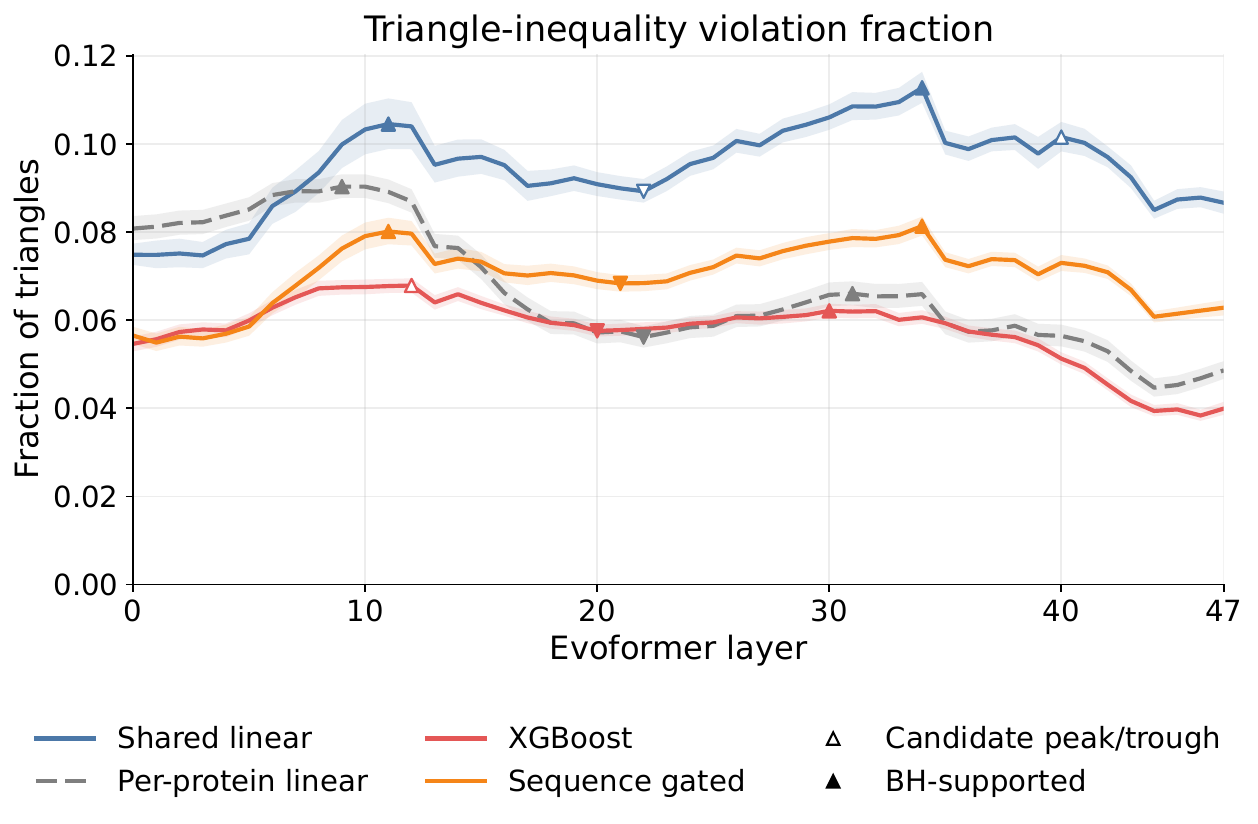}
  \caption{Triangle-inequality violation fraction across Evoformer layers for all unordered residue triples, for the shared linear, per-protein linear, XGBoost, and sequence-gated probes. Curves are equal-weight means over 552 proteins and bands are paired-protein bootstrap percentile 95\% intervals from 2,000 replicates. Hollow markers are shape-qualified candidate peaks (upward) and troughs (downward); filled markers are candidates supported after Benjamini--Hochberg correction (\S\ref{sec:extrema}).}
  \label{fig:supp-triangle-violation-fraction}
\end{figure}

\begin{figure}[t]
  \centering
  \includegraphics[width=0.86\linewidth]{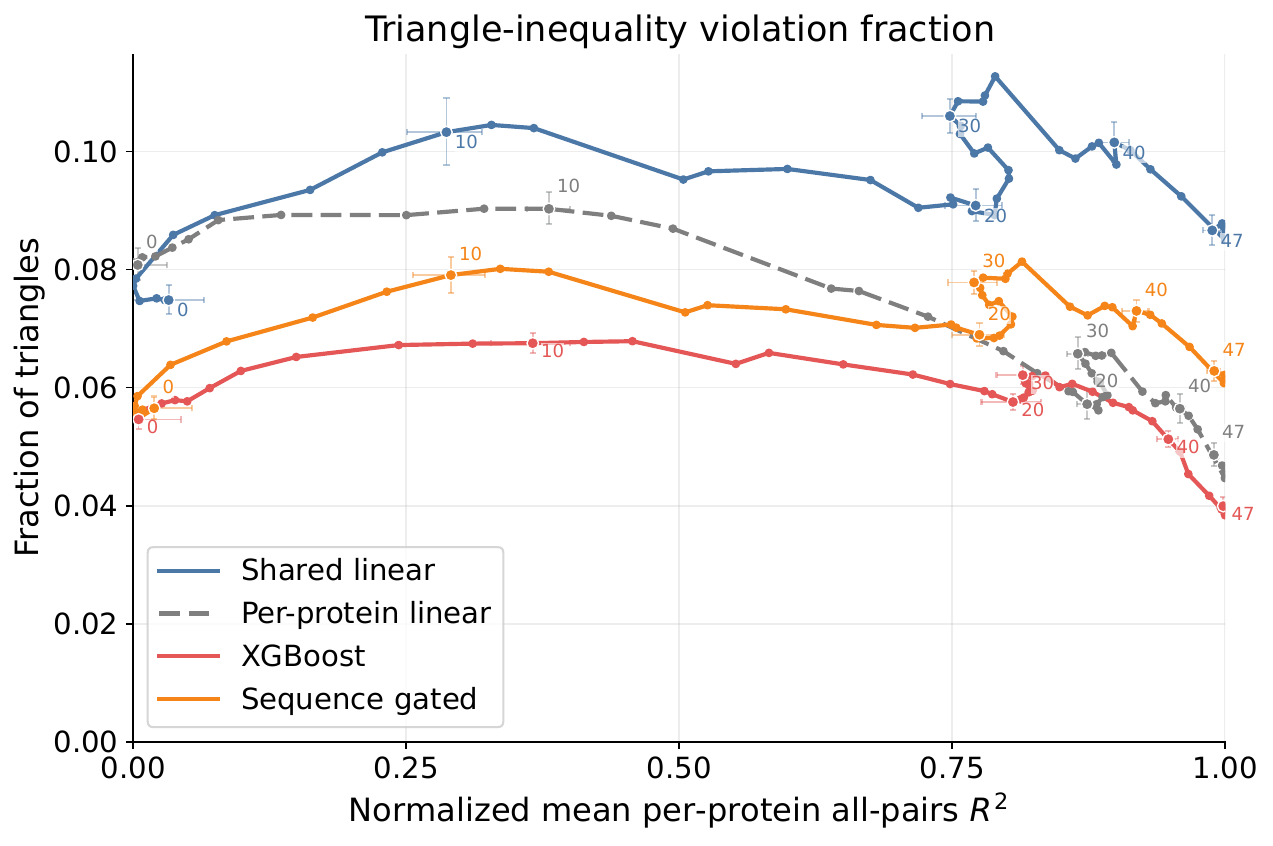}
  \caption{Triangle-inequality violation fraction versus normalized distance decodability for all unordered residue triples. For each probe, normalized decodability at layer $\ell$ is $(\bar R^2_\ell-\min_k \bar R^2_k)/(\max_k \bar R^2_k-\min_k \bar R^2_k)$, where $\bar R^2_\ell$ is the equal-weight mean of per-protein $R^2$ (\S\ref{sec:metrics}). Numbers identify milestone layers 0, 10, 20, 30, 40, and 47; horizontal and vertical bars at those layers are jointly bootstrapped 95\% intervals.}
  \label{fig:supp-triangle-fraction-decodability}
\end{figure}

Figures~\ref{fig:supp-triangle-violation-fraction} and \ref{fig:supp-triangle-fraction-decodability} are the violation-fraction companions to the normalized-mean-violation panels of main-text Figure~2.

\clearpage
\subsection{Cumulative ablation of the triangle mechanisms}

\begin{figure}[t]
  \centering
  \begin{subfigure}[t]{0.49\linewidth}
    \centering
    \includegraphics[width=\linewidth]{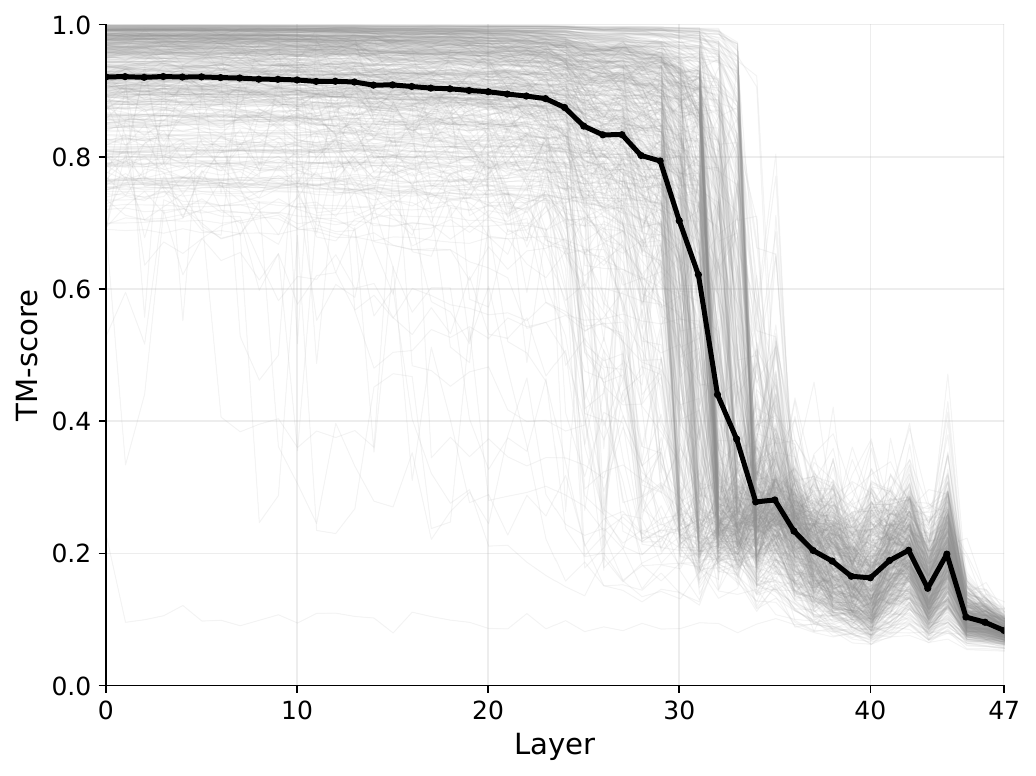}
    \caption{Forward: bypass blocks $0$ through $k$}
    \label{fig:ablate-triangle-attention-tmu-fwd}
  \end{subfigure}\hfill
  \begin{subfigure}[t]{0.49\linewidth}
    \centering
    \includegraphics[width=\linewidth]{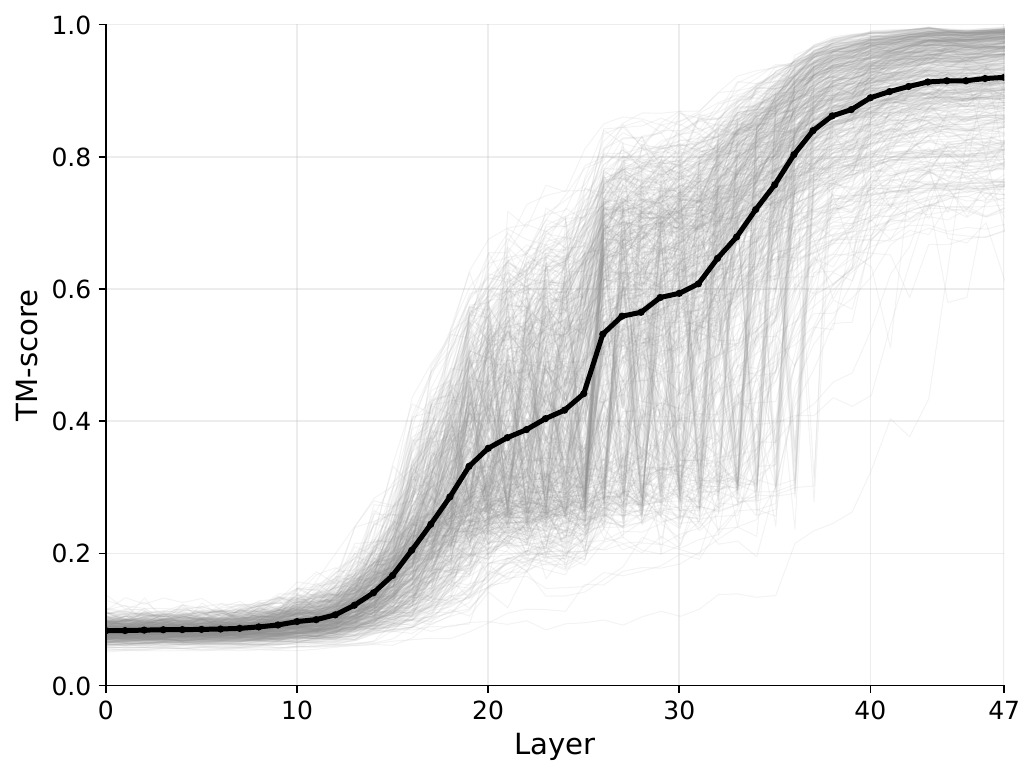}
    \caption{Reverse: bypass blocks $k$ through $47$}
    \label{fig:ablate-triangle-attention-tmu-rev}
  \end{subfigure}
  \caption{Cumulative layerwise ablation of the triangle mechanisms (starting-node Triangle Attention, ending-node Triangle Attention, and Triangle Multiplicative Update bypassed together) on the combined ECOD+CAMEO cohort. In the forward series the modules are bypassed first in layer 0 and then cumulatively through successively later layers; in the reverse series they are bypassed first in layer 47 and then cumulatively through successively earlier layers. TM-score is measured against the experimental structure. Each light gray trace is an individual protein and the black curve is the cohort mean over 552 proteins.}
  \label{fig:supp-triangle-mechanism-ablations}
\end{figure}

Figure~\ref{fig:supp-triangle-mechanism-ablations} shows the joint triangle-mechanism ablation of \S\ref{sec:ablations}; the Pair Transition and MSA Transition ablations are in main-text Figure~4.

\clearpage
\subsection{Accuracy by sequence separation}
\label{sec:strat-acc}

\begin{figure}[t]
  \centering
  \includegraphics[width=\linewidth]{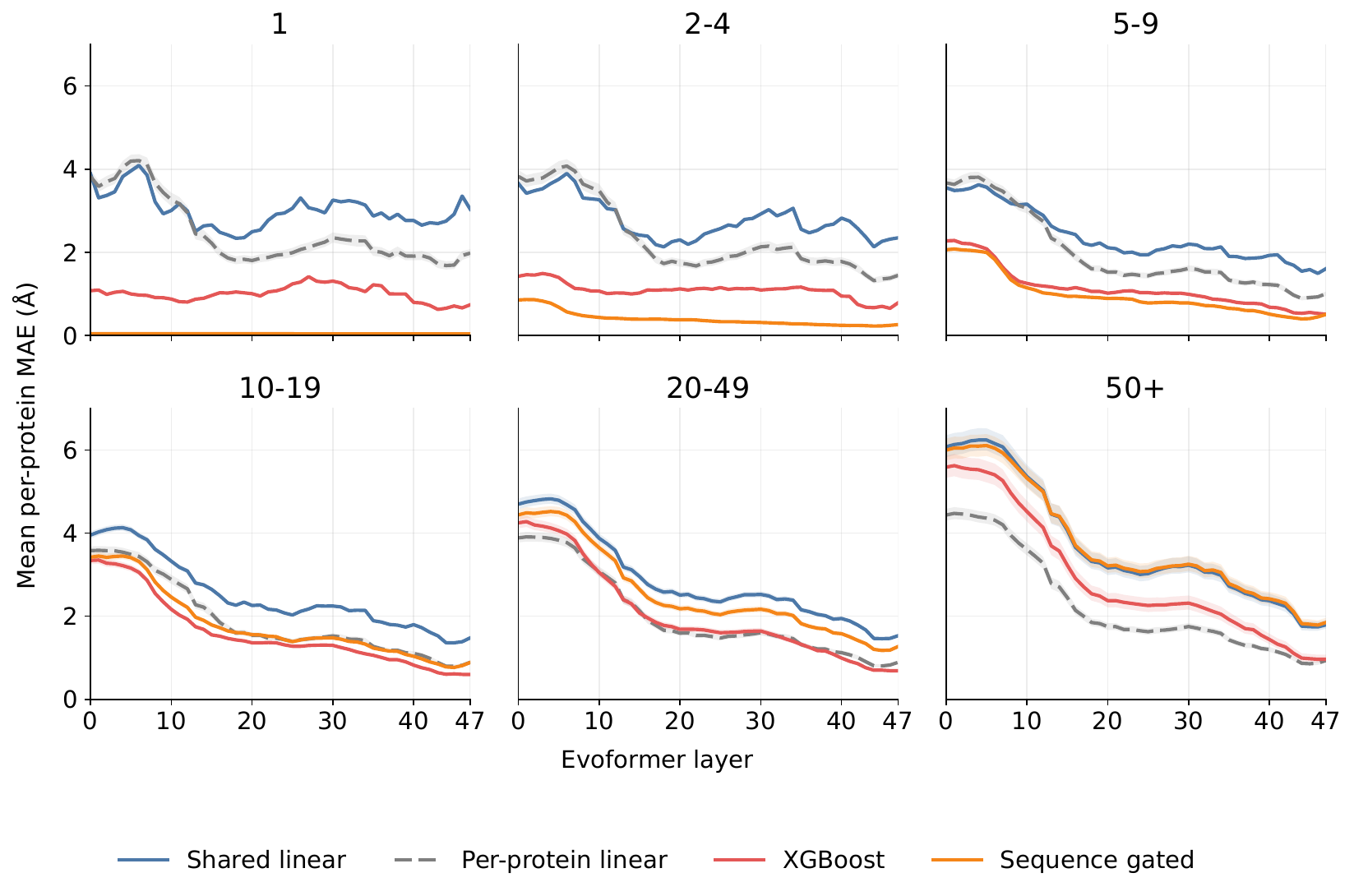}
  \caption{Distance-prediction mean absolute error (MAE) across Evoformer layers, stratified by sequence separation $|i-j|$ into bins of 1, 2--4, 5--9, 10--19, 20--49, and at least 50 residues. Within each protein and bin, MAE is the mean of $|d_{ij}-\hat d_{ij}|$ in \AA; curves average those per-protein values with equal protein weight. Bands are paired-protein bootstrap percentile 95\% intervals from 2,000 replicates. The first five bins contain 552 proteins; the at-least-50 bin contains the 549 proteins with finite support.}
  \label{fig:supp-sequence-separation-mae}
\end{figure}

\subsection{MDS reconstructions across layers}

\begin{figure}[t]
  \centering
  \includegraphics[width=\linewidth]{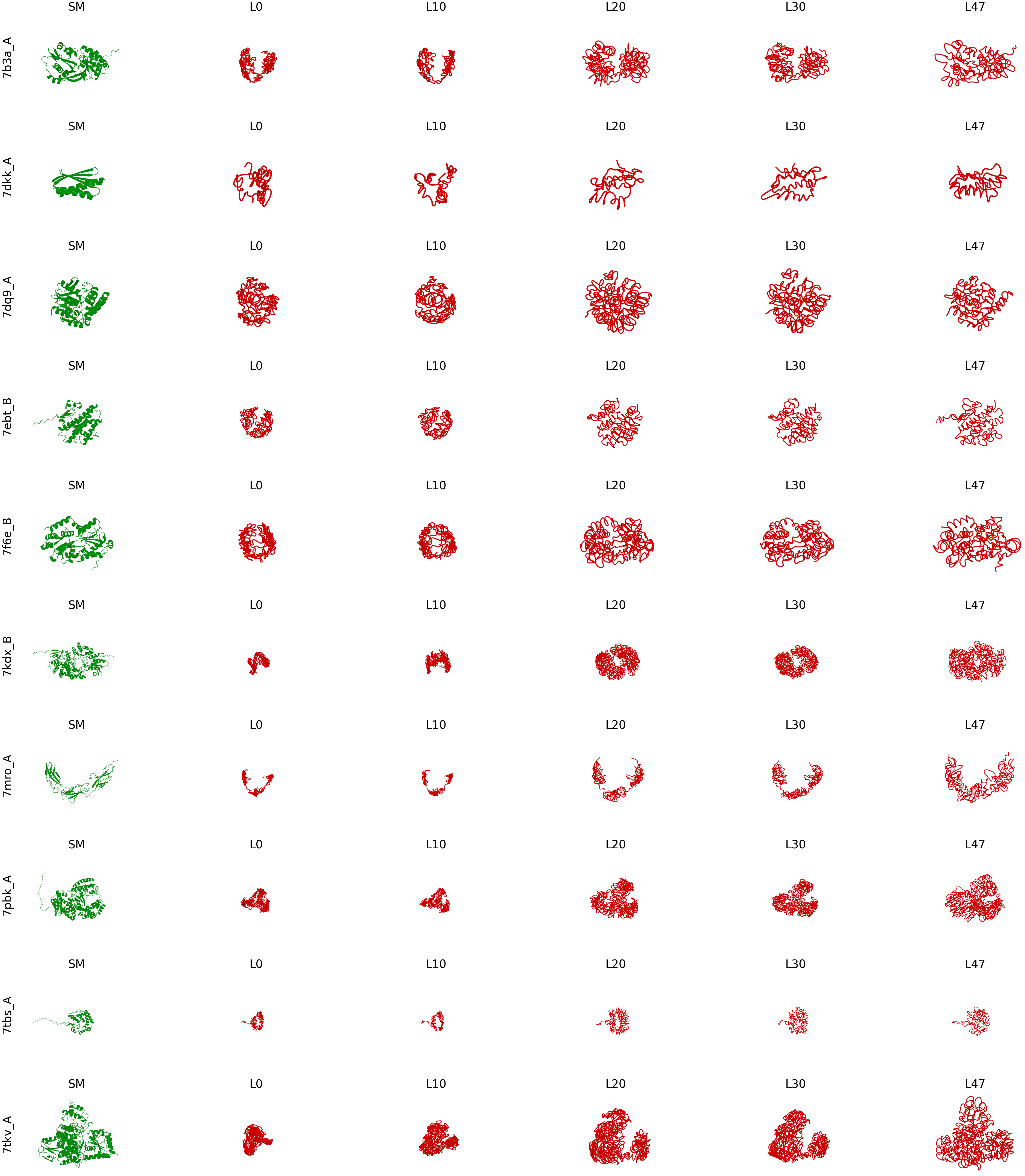}
  \caption{MDS-reconstructed C$\alpha$ traces from held-out-fold shared-OLS distance decoders for ten proteins at Evoformer layers 0, 10, 20, 30, and 47, alongside OpenFold's final Structure Module output (SM). Each row uses a fixed orientation, framing, and scale across panels; framing is independent across proteins. SM cartoons are green and MDS C$\alpha$ traces are red.}
  \label{fig:mds-structures}
\end{figure}

\clearpage
\subsection{Individual-protein trajectories}

\begin{figure}[t]
  \centering
  \includegraphics[width=\linewidth]{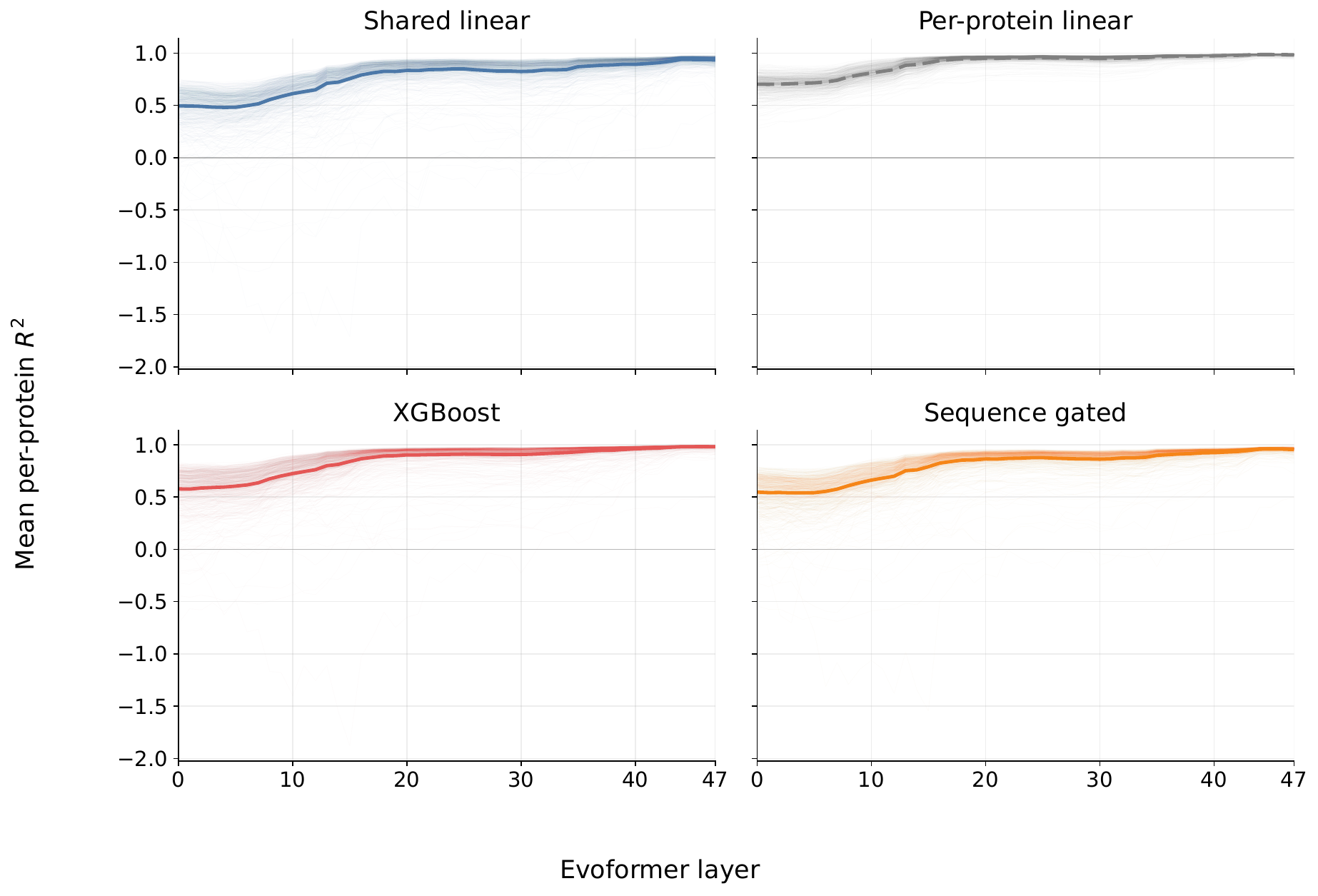}
  \caption{Individual-protein distance-decoding $R^2$ trajectories across Evoformer layers, separated by probe family. For each protein, $R^2 = 1-\sum_{ij}(d_{ij}-\hat d_{ij})^2/\sum_{ij}(d_{ij}-\bar d)^2$ is computed from fivefold out-of-fold predictions.}
  \label{fig:supp-individual-r2}
\end{figure}

\begin{figure}[t]
  \centering
  \includegraphics[width=\linewidth]{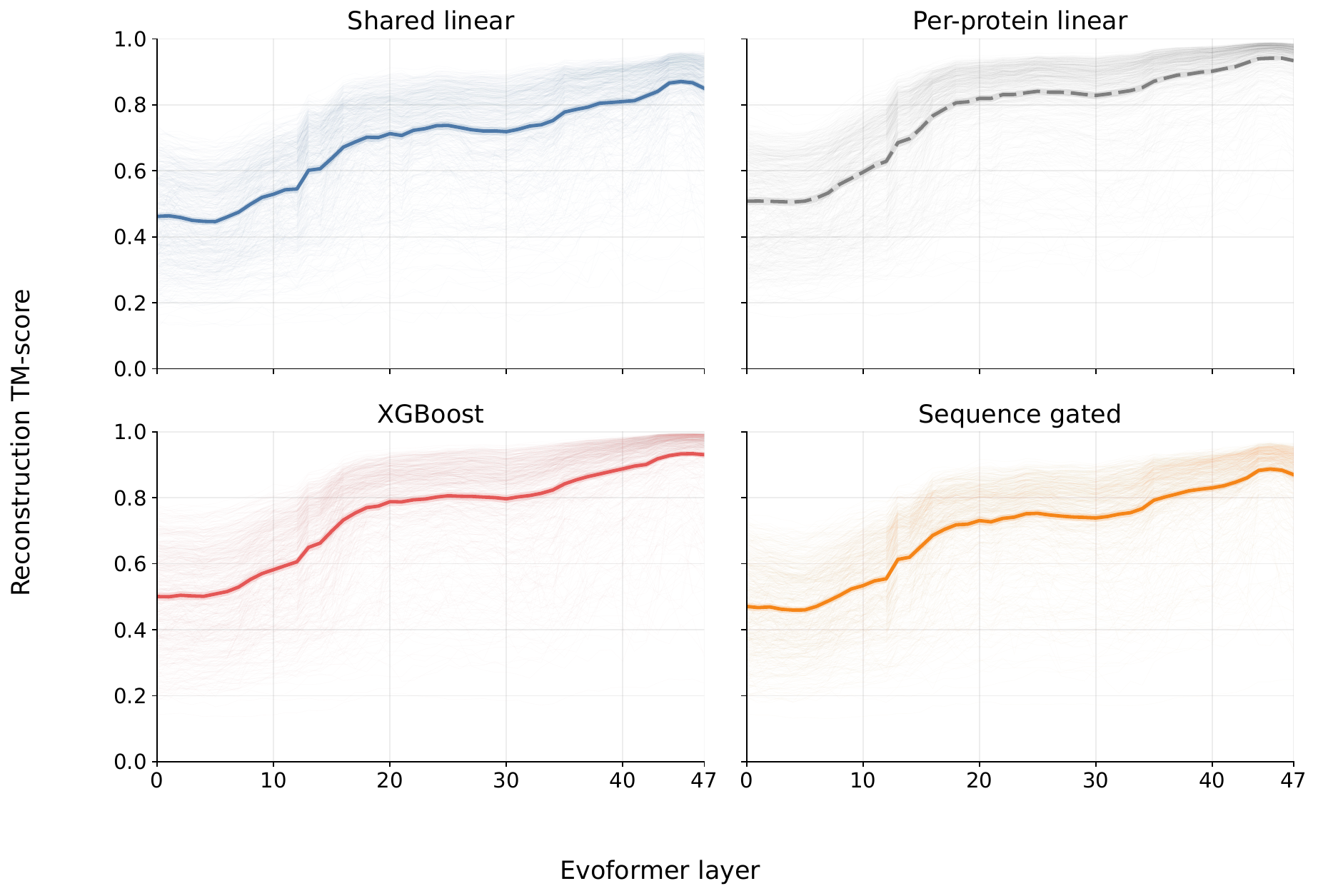}
  \caption{Individual-protein reconstruction TM-score trajectories across Evoformer layers, separated by probe family. At each layer, classical MDS reconstructs three-dimensional C$\alpha$ coordinates from the probe-predicted distance matrix and TM-score compares that reconstruction with OpenFold's final predicted structure (\S\ref{sec:mds}).}
  \label{fig:supp-individual-tm-score}
\end{figure}

\begin{figure}[t]
  \centering
  \includegraphics[width=\linewidth]{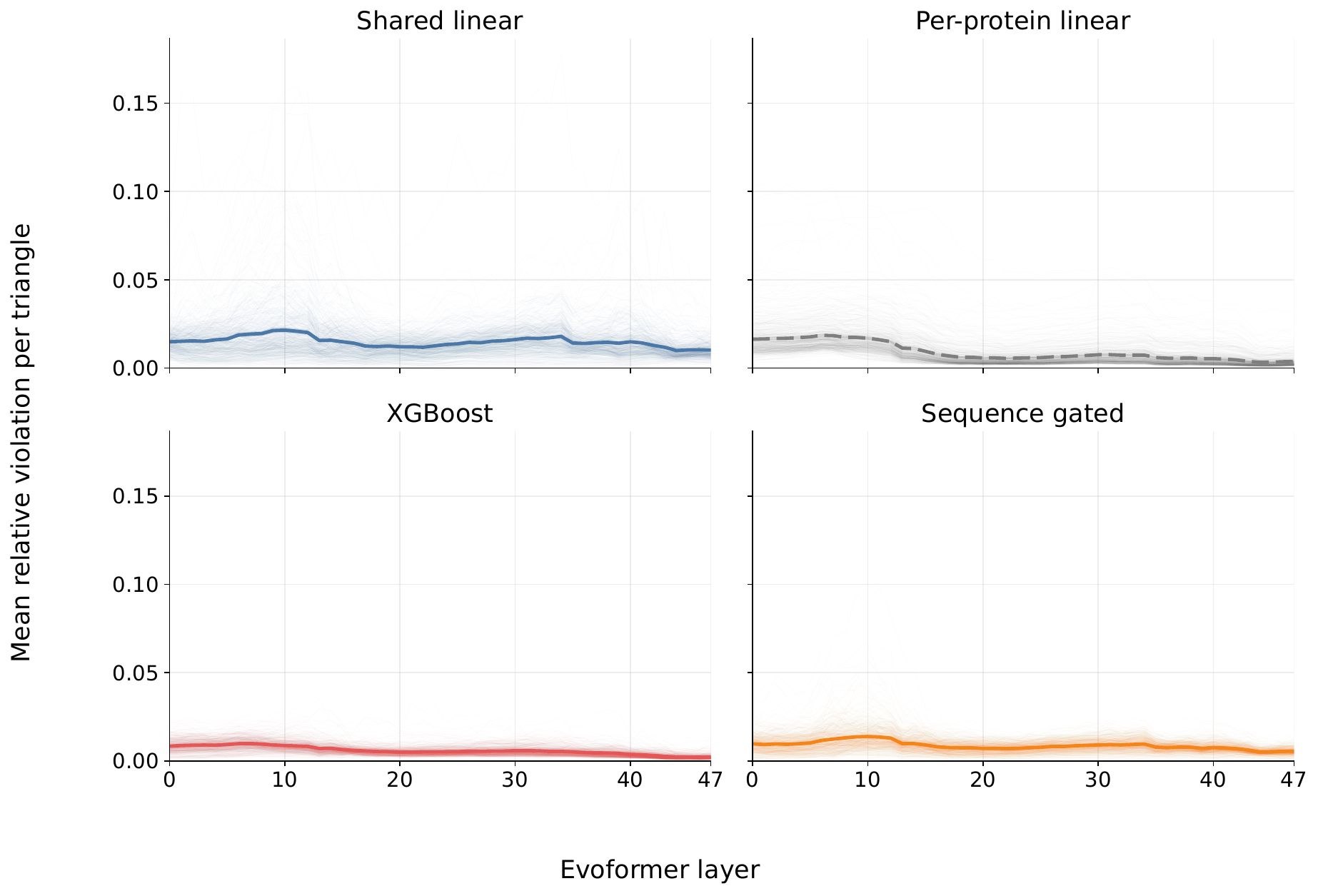}
  \caption{Individual-protein normalized mean triangle-violation trajectories across Evoformer layers, separated by probe family. For each unordered residue triple, $\delta=\max(\hat c-\hat a-\hat b,0)$ for sorted decoded sides, with excess at most $10^{-12}$~\AA{} set to zero; the normalized value is the mean of $\delta/\sqrt{(a^2+b^2+c^2)/3}$ over all triangles, including non-violating triangles, using reference side lengths from OpenFold's final predicted structure (\S\ref{sec:triangle-defs}).}
  \label{fig:supp-individual-normalized-triangle-violation}
\end{figure}

\begin{figure}[t]
  \centering
  \includegraphics[width=\linewidth]{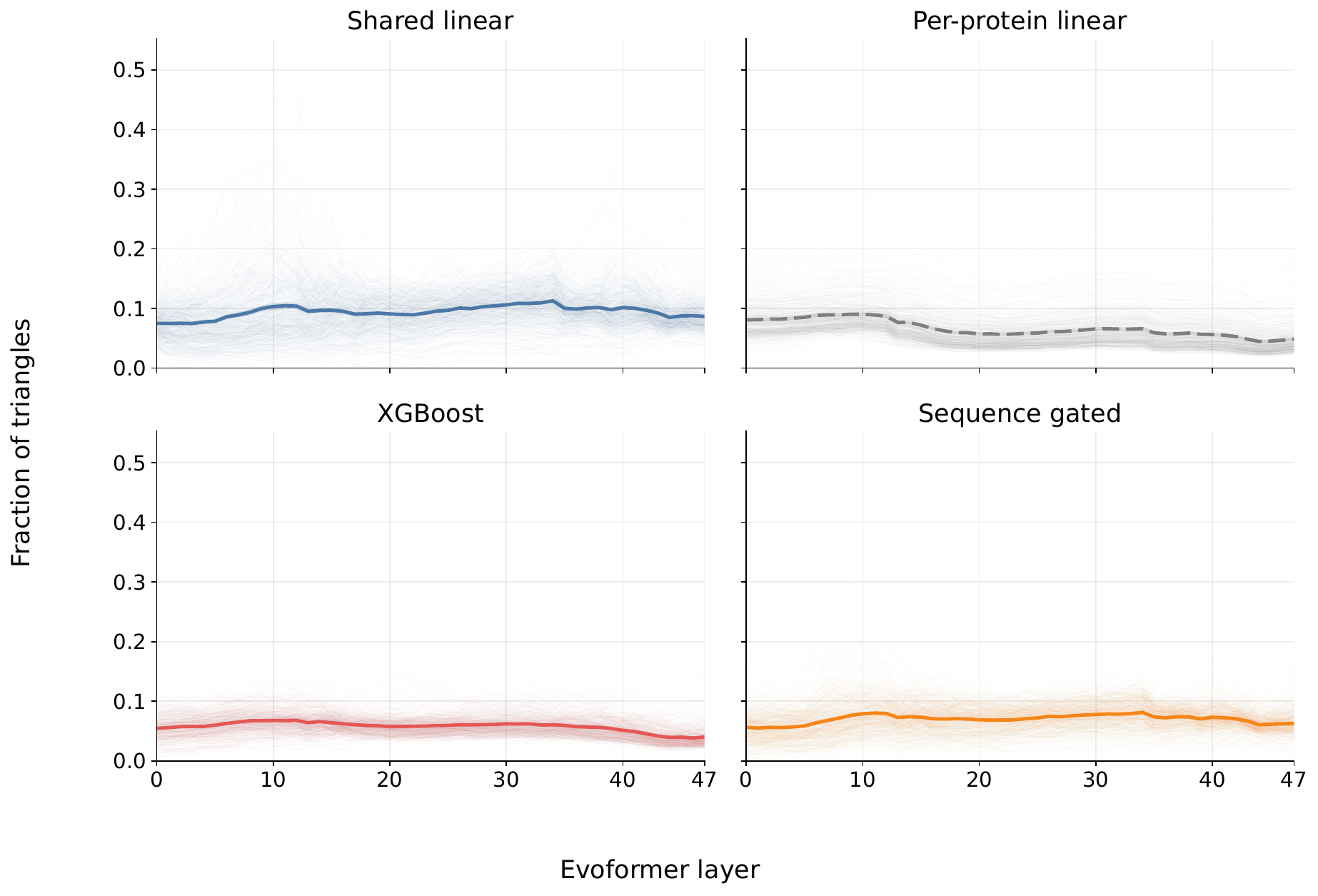}
  \caption{Individual-protein triangle-violation fraction trajectories across Evoformer layers, separated by probe family. The violation fraction is the number of unordered residue triples with $\hat c-\hat a-\hat b > 10^{-12}$~\AA{} divided by the total number of triples in the protein (\S\ref{sec:triangle-defs}).}
  \label{fig:supp-individual-triangle-violation-fraction}
\end{figure}

Figures~\ref{fig:supp-individual-r2}--\ref{fig:supp-individual-triangle-violation-fraction} show the per-protein trajectories underlying the cohort means in the main text, one panel per probe family. Thin colored lines are the 552 individual proteins, the emphasized line is their equal-weight mean, and the band is its paired-protein bootstrap percentile 95\% interval from 2,000 replicates. Shared linear, XGBoost, and sequence-gated probes evaluate unseen proteins; per-protein linear probes evaluate unseen residue pairs within the same protein.

\clearpage
\section{Analysis cohort}
\label{sec:protein-list}

The 154 CAMEO chain identifiers and 400 ECOD domain identifiers drawn for the
cohort of \S\ref{sec:cohort} are listed below.

\paragraph{CAMEO.}
{\small\raggedright\sloppy
\texttt{6tf4\_A 6wmk\_A 6wqc\_A 6zpp\_A 7ab3\_E 7ar0\_B 7atr\_A 7au7\_A 7awk\_A 7b0d\_A
7b1k\_B 7b1w\_F 7b26\_C 7b28\_F 7b29\_A 7b2a\_A 7b2o\_A 7b3a\_A 7b4q\_A 7b7t\_A
7bbz\_A 7bcz\_A 7bew\_A 7bhy\_A 7cjs\_B 7dcm\_A 7dfe\_A 7djy\_A 7dk9\_A 7dkk\_A
7dko\_A 7dmf\_A 7dms\_A 7dnu\_A 7don\_B 7dq9\_A 7dru\_C 7dtp\_A 7dup\_A 7dut\_A
7dvn\_A 7e0m\_A 7e2v\_A 7e37\_A 7e3z\_A 7ebt\_B 7ee3\_C 7ef6\_A 7eqx\_A 7esx\_A
7et8\_A 7ewu\_B 7ezg\_A 7f0h\_A 7f6e\_B 7f8a\_A 7fbh\_A 7fbp\_B 7fe3\_A 7ffa\_A
7fh3\_A 7fip\_A 7jiz\_A 7jt9\_A 7kdx\_B 7kik\_A 7kiu\_A 7kos\_A 7kqv\_D 7ksp\_A
7kua\_A 7kzh\_A 7l6j\_A 7l6y\_A 7l8n\_A 7lbu\_A 7lc5\_A 7lgr\_A 7ljh\_A 7lnu\_B
7lqm\_D 7lqn\_A 7ls0\_B 7lvz\_D 7m4n\_A 7mcc\_A 7mfi\_A 7mfw\_B 7mpz\_A 7mrk\_D
7mro\_A 7mu9\_A 7mwr\_A 7ndr\_A 7nl4\_A 7ntn\_A 7ny6\_A 7o49\_F 7obm\_A 7ofn\_A
7on9\_A 7ool\_A 7ouq\_A 7p3b\_B 7p3t\_B 7p82\_C 7pab\_A 7pbk\_A 7plb\_B 7poh\_B
7ppp\_A 7pq7\_A 7pqf\_A 7prd\_A 7puo\_A 7q03\_A 7q1b\_A 7q47\_A 7r84\_A 7rbw\_A
7rdt\_A 7re4\_A 7re6\_A 7s94\_C 7sf6\_A 7sir\_A 7spo\_C 7spp\_C 7stt\_A 7stz\_C
7swh\_A 7swk\_B 7sy9\_A 7t24\_A 7t5f\_B 7t71\_A 7t9w\_B 7ta9\_A 7tav\_B 7tbs\_A
7tcb\_B 7thw\_A 7tj1\_D 7tkv\_A 7v1v\_A 7v5y\_B 7vdy\_A 7vmu\_A 7vnb\_A 7vpf\_A
7vw0\_B 7w83\_A 7wbr\_A 7wgk\_A}
\par}

\paragraph{ECOD.}
{\small\raggedright\sloppy
\input{eocd-protein-list}
\par}

%% file: eocd-protein-list.tex
\texttt{000002954
000003088
000002873
000002527
001030873
001688250
002034390
000002523
000002641
000002570
000002798
000002994
000002677
002084847
000002973
000003374
000003017
002323964
000002719
000002735
001779586
000003566
000002581
000002843
001117697
000136907
000004005
000003887
000004074
000004062
000004064
000004008
000004094
000004091
000003705
000004083
000003934
002593330
000003933
000003958
000003944
000003992
000003951
000003965
000003974
000003969
000003979
000004113
001097701
000003977
000004887
000185466
000004879
001291927
000118923
001830962
000185467
001145805
001807453
000004884
000004886
000004893
000004875
000004781
002084829
000004782
000004867
000004868
001498254
001070254
001885546
000004796
001720175
000004785
001310594
000004627
000146396
000004614
000004595
001108139
001447889
000004447
000004519
000004518
000004609
000004620
000004619
002323751
000184628
000004480
000004499
001117739
002089786
001826907
001291945
000004605
001806534
001673889
001148163
001291870
000000603
000000498
001691936
000000590
000000516
001314520
000000479
001298733
000000544
000000187
000000033
000000160
000146640
000000362
000000359
001291989
000000515
000000299
000000707
000000087
000000300
000000922
000000711
001562394
000000739
000000024
000142351
001318160
000138885
000000050
001714442
000000269
000000565
000000976
000000929
000000022
001088780
001349406
000000464
000000967
000000171
001088763
000000309
001568669
002321120
000000233
000000888
001298163
000196893
000000238
000001773
001148147
000001981
001108152
000001831
000002063
000001575
000001712
000001967
002324052
000001694
000001555
000001449
001097183
000001542
000001670
000001385
000001532
002324091
001031014
000002060
000001682
000001527
001828335
001003950
000001920
000001431
000001782
000146284
000144872
000001858
000001181
000001881
000185334
000002058
000001330
000001869
000185426
000001337
000001166
000001300
001839918
000001219
000001308
001693512
000002069
000002028
001405130
000002023
002082647
000008982
000008757
001145693
000008735
000008683
001954222
000008740
000008901
000008756
000008684
000008768
000008959
000008935
000008969
001148158
000008794
000008775
000008897
000008869
000008910
000008930
001312438
000008947
002387870
000008949
001828332
000008770
000008699
000008762
000008710
000008922
000008772
001498264
002323965
000999472
002387784
000008662
000008857
000008886
001722659
000008843
000008876
000146885
000008810
000008798
000008654
000008822
000008863
000008666
001840971
001518739
001518804
001562402
001518843
000010467
000009925
001411388
000146903
000009156
000010881
000009774
000011036
002482206
000010984
000010918
000009308
000009183
002323904
000010986
000146896
000010946
001031067
000009692
001681437
001148170
002095512
000010829
000185903
000010830
001291953
001082810
000009694
000009671
000009344
000009166
001889963
001889960
000010869
001295875
001002413
000009583
000185598
001124216
000009610
001291931
001843702
000009578
000009122
000010801
000009038
000007572
000007326
001420603
001295853
000007419
001891408
001498209
000007351
000185622
000007422
000007385
001447923
000007109
000007435
000185064
000007356
000007412
000007128
000185718
000007075
000007376
001108083
000007068
000146397
000119296
000007869
000007830
000007872
000007811
000007839
000007812
000007821
000007890
000007820
001870507
002095500
000007835
001866734
000007834
000007822
000007836
000007900
001563534
000134473
001563549
000007873
000007855
000007831
001107994
001885529
000011306
000011305
001147788
000011290
001030983
000011286
001806466
000011316
000006870
000006823
001563539
000011338
001563542
001567475
000006816
000011098
000011060
000006833
000011054
000011108
000011067
002576225
000006814
001002434
000223959
000005925
000006206
000005265
000006155
001548153
001839911
000006035
000005488
000006203
000005100
001121976
000006492
000005098
000005171
000005190
000005849
000006001
000224375
000006395
000006560
000006238
000184879
000005344
001411401
000146378
}